\documentclass{article}

\usepackage{amsmath,amsfonts,bm}

\def\eqref#1{equation~\ref{#1}}

\def\1{\bm{1}}

\DeclareMathAlphabet{\mathsfit}{\encodingdefault}{\sfdefault}{m}{sl}
\SetMathAlphabet{\mathsfit}{bold}{\encodingdefault}{\sfdefault}{bx}{n}

\usepackage[svgnames]{xcolor}

\newcommand{\sref}[1]{\S\ref{#1}} %
\newcommand{\sealp}{SEAL-pose}

\usepackage{microtype}
\usepackage{graphicx}
\usepackage{subcaption}
\usepackage{algorithm}
\usepackage{algorithmic}
\usepackage{siunitx}
\usepackage{float}
\usepackage{multirow}
\usepackage{booktabs} 
\usepackage{hyperref}

\usepackage[preprint]{icml2026}

\usepackage{amsmath}
\usepackage{amssymb}
\usepackage{mathtools}
\usepackage{amsthm}

\usepackage[capitalize,noabbrev]{cleveref}

\theoremstyle{plain}

\theoremstyle{definition}

\theoremstyle{remark}

\usepackage[textsize=tiny]{todonotes}

\icmltitlerunning{SEAL-pose: Enhancing 3D Human Pose Estimation via a Learned Loss for Structural Consistency}

\begin{document}

\twocolumn[
  \icmltitle{SEAL-pose: Enhancing 3D Human Pose Estimation \\via a Learned Loss for Structural Consistency}



  \icmlsetsymbol{equal}{*}

  \begin{icmlauthorlist}
    \icmlauthor{Yeonsung Kim}{snu,equal}
    \icmlauthor{Junggeun Do}{TAMU,equal}
    \icmlauthor{Seunguk Do}{snu}
    \icmlauthor{Sangmin Kim}{snu}
    \icmlauthor{Jaesik Park}{snu}
    \icmlauthor{Jay-Yoon Lee}{snu}
    
  \end{icmlauthorlist}

  \icmlaffiliation{snu}{Seoul National University, South Korea}
  \icmlaffiliation{TAMU}{Texas A\&M University, College Station, Texas, USA}
  
  \icmlcorrespondingauthor{Jaesik Park}{jaesik.park@snu.ac.kr}
  \icmlcorrespondingauthor{Jay-Yoon Lee}{lee.jayyoon@snu.ac.kr}

  \icmlkeywords{Machine Learning, ICML}

  \vskip 0.3in
]



\printAffiliationsAndNotice{\icmlEqualContribution}

\begin{abstract}
3D human pose estimation (HPE) is characterized by intricate local and global dependencies among joints. Conventional supervised losses are limited in capturing these correlations because they treat each joint independently. Previous studies have attempted to promote structural consistency through manually designed priors or rule-based constraints; however, these approaches typically require manual specification and are often non-differentiable, limiting their use as end-to-end training objectives. We propose \sealp, a data-driven framework in which a learnable loss-net trains a pose-net by evaluating structural plausibility. Rather than relying on hand-crafted priors, our joint-graph-based design enables the loss-net to learn complex structural dependencies directly from data. Extensive experiments on three 3D HPE benchmarks with eight backbones show that \sealp~reduces per-joint errors and improves pose plausibility compared with the corresponding backbones across all settings. Beyond improving each backbone, \sealp~also outperforms models with explicit structural constraints, despite not enforcing any such constraints. Finally, we analyze the relationship between the loss-net and structural consistency, and evaluate \sealp~in cross-dataset and in-the-wild settings.
\end{abstract}

\begin{figure}[t]
    \centering
    \includegraphics[width=0.95\linewidth]{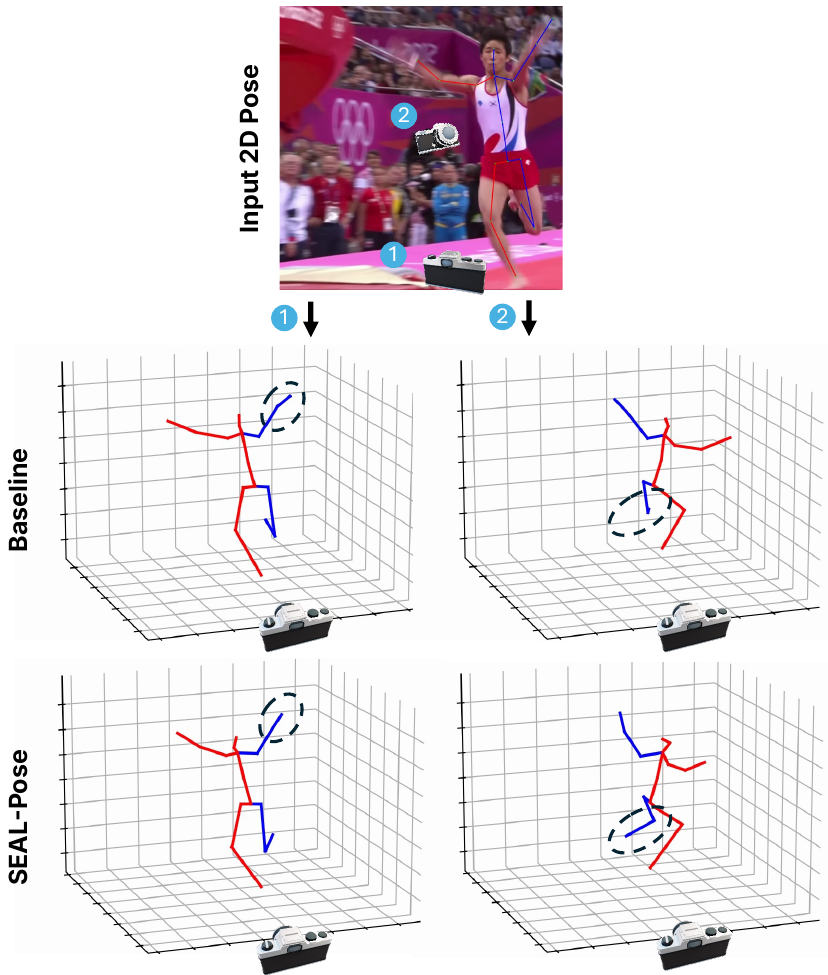}
    \caption{\sealp~improves structural consistency by preserving skeletal symmetry and kinematic connectivity; under the same 2D input, KTPFormer (baseline) can exhibit local joint inconsistencies (e.g., ankle/wrist) that propagate to the overall limb configuration.}
    \label{fig:teasure}
\end{figure}

\section{Introduction}

{3D human pose estimation (3D HPE) requires predicting accurate joint positions while preserving the underlying anatomical structure~\citep{survey1}. This task is particularly difficult because the output space is governed by complex local and global dependencies between joints. However, common training objectives such as mean squared error (MSE) and mean per-joint position error (MPJPE) penalize individual joint errors without accounting for structural consistency, which often results in implausible or anatomically inconsistent poses. Therefore, it is critical to effectively model the structures in the output space to predict accurate and plausible 3D poses. Previous studies~\citep{jran, limbpose, posegrammar, posegrammar2, kim2024toward} have attempted to capture such structural dependencies using hand-crafted constraints, which can fail to reflect the natural structural consistency of human poses.

To overcome these limitations, we propose \sealp, a framework that provides structural guidance for 3D pose estimation via a trainable loss. 
Building on Structured Energy As Loss (SEAL)~\citep{seal}, we adapt learnable energy-based loss to 3D human pose estimation, where the output is a high-dimensional 3D geometry constrained by skeletal topology. 
At the core of \sealp\ is a skeleton-aware loss-net jointly optimized with a pose estimation model (pose-net). Leveraging skeletal topology as an inductive bias, the loss network learns local and global structural relations—from adjacent joints to long-range dependencies—directly from data.
We construct joint-wise coupled 2D–3D inputs for the loss-net, providing observation-conditioned signals that it can aggregate across joints to capture structural plausibility beyond per-joint regression. Although alternating optimization between the pose-net and loss-net can seem to be brittle, a simple greedy hyperparameter tuning shows a reliable convergence of \sealp~in practice.

Finally, to evaluate structural quality beyond standard errors such as MPJPE, we introduce two complementary metrics—Limb Symmetry Error (LSE) and Body Segment Length Error (BSLE)—that quantify symmetry and segment-length consistency of predicted poses.
Our framework is model-agnostic and easily extends to single-frame, multi-frame, and diffusion-based pose estimation models by simply incorporating the loss network during training.
Extensive experiments on Human3.6M~\citep{h36m_pami}, MPI-INF-3DHP~\citep{3dhp}, and Human3.6M 3D WholeBody~\citep{h3wb} demonstrate that \sealp~not only reduces per-joint errors (Table \ref{tab:h36m}, \ref{tab:3dhp}, \ref{tab:h3wb}) but also produces more plausible poses, evaluated by the newly proposed LSE and BSLE (Table~\ref{tab:metric} and Figure~\ref{fig:struct-err-3dhp},~\ref{fig:gbi-3dhp}). 
In summary, the contributions of this paper are as follows:
\begin{itemize}
  \item[(i)] We redesign SEAL for 3D pose estimation and propose \sealp~to bridge the gap between previous discrete label–space dependency modeling and continuous 3D geometric outputs governed by skeletal constraints.
  \item[(ii)] We introduce a skeleton-aware graph loss-net with joint-wise 2D-3D coupled representations, and demonstrate consistent improvements across datasets and backbones with no test-time overhead.
  \item[(iii)] We propose structure-aware evaluation metrics (LSE/BSLE) and further explore leveraging diverse hypotheses as negative samples, which provides additional gains in both single-frame and multi-frame settings.
\end{itemize}


\section{Related Work}

\subsection{3D Human Pose Estimation}

3D human pose estimation is a well-established computer vision task involving the prediction of 3D joint positions from 2D images or videos. This task is inherently challenging because it requires inferring spatial relationships while ensuring anatomical plausibility from incomplete visual information. Current approaches generally follow two paradigms: (1) directly predicting 3D poses from images~\citep{coarsetofine,ordinal} or (2) estimating 2D poses first and then lifting them to 3D space~\citep{survey2, survey1}. The 2D-to-3D lifting has been widely adopted due to progress in 2D human pose estimation~\citep{survey2}.

While single frame models such as SimpleBaseline~\citep{simplebaseline} and SemGCN~\citep{semgcn} perform well, they process frames independently and thus cannot enforce temporal consistency, often resulting in jitter. This limitation has motivated multi-frame models that leverage temporal dynamics for more stable and robust 3D pose estimation.

\textbf{Transformer-based models.}
Transformers~\citep{vaswani2023attentionneed} have become a major backbone for 3D human pose estimation. PoseFormer~\citep{poseformer} was the first transformer-based architecture for this task, and PoseFormerV2~\citep{poseformerv2} improved efficiency via frequency-domain representations.
While PoseFormer-style models typically handle spatial relations before temporal modeling, MixSTE~\citep{mixste} interleaves the two through a mixed spatio-temporal encoder.
P-STMO~\citep{pstmo} further introduces a pre-trained spatio-temporal masked autoencoder to capture motion dynamics efficiently.
To incorporate skeletal priors into self-attention, MotionAGFormer~\citep{Mehraban_2024_WACV} combines GCNs with Transformers, whereas KTPFormer~\citep{peng2024ktpformer} injects kinematic and trajectory priors to stabilize attention.

\textbf{Graph-based models.} Graph-based models, such as Graph Convolutional Networks~\citep{gcn} and Graph Attention Networks~\citep{gat}, are widely used in human pose estimation because they naturally encode skeletal structure~\citep{semgcn}, but their local receptive fields limit long-range reasoning. Graformer addresses this by injecting joint and edge-aware priors into the attention mechanism, enabling global, structure-aware interactions~\citep{GraFormer}. Building on this idea, we design a structure-aware loss-net that guides the pose-net to learn the human kinematic structure consistently. 
\begin{figure*}[t]
    \centering
    \includegraphics[width=1\linewidth]{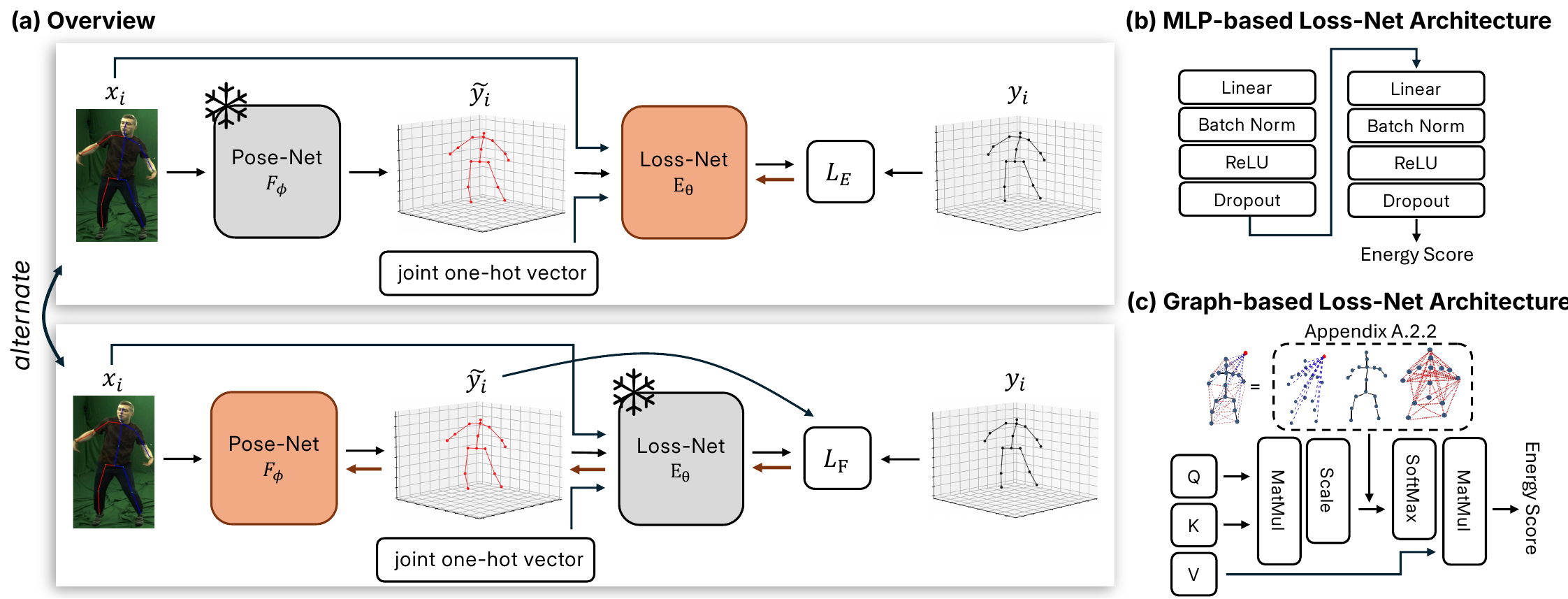}
    \caption{\textbf{(a) Overview of \sealp.} 
    \sealp~combines a pose-net $F_\phi$ that lifts 2D keypoints $x_i$ to a predicted 3D keypoints $\tilde{y}_i$ with a loss-net $E_\theta$ (graph-based or MLP) that predicts an energy score, where $y_i$ denotes the ground-truth 3D pose.  It adopts \textbf{alternating optimization}: it first freezes $E_{\theta}$ and optimizes $F_{\phi}$ to minimize the energy of predicted poses. Then, it freezes $F_{\phi}$ and trains $E_{\theta}$ so that its energy better reflects pose quality.  \textbf{(b-c) Loss-net Architecture Variants.}}
    \label{fig:SEALpose_framework}
\end{figure*}
\subsection{Output Structure of 3D HPE}

3D HPE has inherent challenges such as ambiguity due to incomplete information, which is further compounded in single-frame scenarios. 
To address this issue, previous works have designed methods that consider prior knowledge of human body structure and joint relationships~\citep{posegrammar, posegrammar2, jran, anatomy-aware}, and enforced structural plausibility by explicitly constraining bone length, angles, and symmetry~\citep{lbone, limbpose, anatomical-constraints, anatomy-aware}.

Beyond these approaches, recent studies have explored generating multiple hypotheses or plausible 3D poses to alleviate depth ambiguity and structural uncertainty. For instance,~\cite{kim2024toward} proposed a Biomechanical Pose Generator to augment training data with biomechanically valid poses, along with Binary Depth Coordinates to resolve the depth ambiguity by classifying the joint depths as front or back. Similarly,~\cite{manipose} introduced ManiPose, a manifold-constrained multi-hypothesis approach that estimates the plausibility of each candidate and restricts them to the human pose manifold.

Despite their contributions, most existing methods rely on prior knowledge or predefined rules, which may limit scalability and adaptability. In contrast, we aim to address these limitations by providing a more flexible and general approach for 3D HPE that captures joint dependencies without explicit prior knowledge. In addition, our method, \sealp, is agnostic to model architecture and can potentially be extended to various tasks with complex output structures.
\section{Methodology}
\subsection{Preliminaries: Structured Energy as Loss (SEAL)}
The Structured Energy As Loss (SEAL) framework improves structured prediction by using a structured energy network as a trainable loss. Concretely, SEAL introduces a secondary network (loss-net) that scores the plausibility of a model’s outputs, and uses this score to provide learning signals for the primary predictor (task-net). This formulation encourages the task-net to account for dependencies among output variables without relying on handcrafted constraints.

SEAL is commonly instantiated in two variants: SEAL-static and SEAL-dynamic. SEAL-static uses a fixed, pre-trained loss-net during training, while SEAL-dynamic updates the loss-net jointly as the pose-net evolves. Prior work reports that SEAL-dynamic often yields stronger supervision than SEAL-static. Accordingly, we adopt the SEAL-dynamic formulation in our framework.
\subsection{Designing SEAL for 3D HPE.}
Unlike the discrete structured output spaces typically considered in SEAL, 3D human pose estimation operates in a continuous, high-dimensional coordinate space. In this setting, structure is expressed as geometric relationships on the human skeleton. To make a trainable loss effective in this setting, we design \sealp~around three choices.
First, we implement the loss-net as a skeleton-aware, graph-based model to capture both local and global joint relationships.
Second, we design and evaluate loss-net input representations to tackle the non-trivial challenge of learning compatibility between 2D and 3D skeletal points in a continuous geometric space.
Third, we introduce synthetic negatives to strengthen the loss net’s training signal in deterministic 3D HPE models, which lack the natural diversity of negatives available in probabilistic formulations considered in~\citet{seal}.

\subsection{Training Procedure.}
Our framework consists of two components: (i) pose-net, which can be any 3D HPE model that predicts 3D joint positions from 2D inputs, and (ii) loss-net, a trainable loss function that dynamically learns to evaluate the structural plausibility of predicted poses.
The pose-net and loss-net are updated in an alternating manner, allowing the loss-net to adapt to the evolving predictions of the pose-net and provide informative structural learning signals. Specifically, the pose-net is optimized with a combined objective consisting of the standard mean squared error and the loss-net energy score, which we denote as $\mathcal{L}_F$, while the loss-net is trained with an energy-shaping objective $\mathcal{L}_E$ that assigns lower energy to ground-truth poses and higher energy to implausible predictions. Although alternating updates can be unstable, we find that combining the energy term with supervised loss yields stable training and often faster convergence (\sref{subsec:trian_stability}).

Our framework can be seamlessly combined with various backbone models, from single-frame lifting models to multi-frame models, since the loss-net can be introduced independently of the pose-net architecture. Furthermore, it does not incur any additional inference cost because the loss-net is only used during training.
The overall procedure is summarized in Algorithm~\ref{alg:sealp}, and detailed explanations are provided in Appendix~\ref{SEAL-pose-detail}.

\begin{algorithm}[t]
\caption{\sealp{} Algorithm}
\label{alg:sealp}
\footnotesize
\begin{algorithmic}[1]
\REQUIRE $(\mathbf{x}, \mathbf{y})$: training data (2D inputs and 3D ground-truth outputs)
\REQUIRE $F_\phi$: pose-net with parameters $\phi$
\REQUIRE $E_\theta$: loss-net with parameters $\theta$
\REQUIRE \texttt{optimizer}$_\phi$: optimizer for pose-net
\REQUIRE \texttt{optimizer}$_\theta$: optimizer for loss-net
\REQUIRE $T$: number of training iterations

\STATE Initialize $\phi_0$, $\theta_0$ randomly
\FOR{$t = 1$ {\bf to} $T$}
  \STATE Sample mini-batch $B_t = \{(x_i, y_i)\}_{i=1}^{|B_t|}$
  \STATE $\tilde{y}_i \gets F_{\phi_{t-1}}(x_i)$ for all $(x_i,y_i)\in B_t$
  \STATE $g_\theta \gets \nabla_\theta \frac{1}{|B_t|}\sum_{(x_i,y_i)\in B_t}
         L_E(x_i, y_i, \tilde{y}_i;\theta_{t-1})$
  \STATE $\theta_t \gets \theta_{t-1} - \eta_\theta\, g_\theta$
  \STATE $g_\phi \gets \nabla_\phi \frac{1}{|B_t|}\sum_{(x_i,y_i)\in B_t}
         L_F(x_i, y_i;\theta_t)$
  \STATE $\phi_t \gets \phi_{t-1} - \eta_\phi\, g_\phi$
\ENDFOR
\end{algorithmic}
\end{algorithm}

\subsection{Graph-based Loss-Net.}
In \sealp, the capability of the loss-net to capture structural dependencies in the output space is crucial for guiding the pose-net toward more plausible predictions. Beyond merely modeling local (short-range) and global (long-range) relations, the loss-net must aggregate them into a unified, whole-pose signal that can effectively guide the task network.
To further enhance this ability, we use a graph-based design for the loss-net, enabling more effective use of skeletal structure.
We adopt Graphormer~\citep{ying2021transformers} as the loss-net backbone with task-specific simplifications, using self-attention to model both short- and long-range joint dependencies.
This design results in a more expressive and structure-aware trainable loss function, providing stronger structural guidance for the task network and ultimately yielding more consistent and coherent 3D pose predictions. 

In addition, to assess the suitability of graph-based structures, we also implement an MLP-based loss-net as a baseline. This comparison allows us to examine whether utilizing graph structure provides benefits over a simpler fully-connected neural network. Overall implementation details of loss-nets are provided in Appendix~\ref{appendix:graphormer}.

\subsection{Loss-net Input Design via Early Fusion}
In 3D HPE, the output space is continuous, with infinitely many pose candidates. As a result, subtle differences in 2D cues and 3D hypotheses are often hard to disentangle in representation space when the two modalities are weakly coupled. Therefore, the energy should be conditioned on the 2D observation—ideally via early fusion—so that it reflects how compatible a 3D hypothesis is with the given evidence.

In contrast, SEAL was originally motivated by discrete structured prediction, where the loss-net typically operates on label embeddings. Its input design does not explicitly model 2D-conditioned 3D compatibility in geometric space. To better reflect this requirement in 3D HPE, we design a joint-aligned early-fusion input that directly exposes 2D--3D compatibility signals to the loss-net. Combined with a graph-based loss-net, early fusion captures fine-grained joint-wise 2D–3D compatibility, and message passing lifts these local cues to whole-skeleton consistency, producing globally coherent energies. Details are provided in Appendix~\ref{appendix:rep}.

\subsection{Negative Sampling Strategies}
\label{sec:synthetic-negatives}
Because 3D poses lie in a continuous, high-dimensional space, negatives are effectively unbounded, which can make learning unstable.
We therefore use a margin-based objective with hard negatives to provide a stable separation signal and enforce meaningful energy gaps.

\textbf{Hard negatives from diffusion samples.}
A diffusion-based model generates multiple candidate poses per input. We construct a hard negative by selecting the candidate that best matches the 2D observation under reprojection:
\[
\tilde{\mathbf y}_{\text{neg}}
=\arg\min_{k=1,\dots,K}\mathrm{MPJPE}_{2D}\!\left(\Pi\!\left(\tilde{\mathbf y}^{(k)}\right),\mathbf u\right),
\]
where $\mathbf u$ denotes the ground-truth 2D keypoints and $\Pi(\cdot)$ is the camera projection. 
This 2D-based selection yields a reprojection-consistent negative that better reflects 2D--3D compatibility.

\textbf{Single-frame negatives from 2D perturbations.}
We construct $K$ negative samples by perturbing the input 2D keypoints and passing them through the same deterministic pose-net:
\begin{align}
x_{\mathrm{noisy}}^{(k)} &= x + \epsilon^{(k)}, \qquad 
\epsilon^{(k)} \sim \mathcal{U}\!\bigl(\|\epsilon\|_2 \le R_{\max}\bigr), \\
\tilde{y}_{\mathrm{neg}}^{(k)} &= F_{\phi}\!\left(x_{\mathrm{noisy}}^{(k)}\right), \qquad k=1,\dots,K.
\end{align}
We keep the original prediction-to-ground-truth margin loss and add a negative--negative ordering regularizer: we sort negatives by their 2D MPJPE and compare adjacent and randomly sampled pairs, encouraging the loss-net energy to capture fine-grained structural discrepancies.
\\
\begin{equation}
\ell_{\text{pair}}(i,j)=\Bigl[\kappa\,\mathrm{MPJPE}\!\bigl(\tilde{y}_{\text{neg}}^{(i)},\tilde{y}_{\text{neg}}^{(j)}\bigr)-E_{\text{neg}}^{(i)}+E_{\text{neg}}^{(j)}\Bigr]_+ .
\end{equation}

\textbf{Multi-frame extension via neighboring frames.}
In the multi-frame setting, we use the same pairwise loss, but leverage neighboring frames to form implicit negative pairs, using window-averaged poses and energies. We sample two nearby start indices $t$ and $s$ and form window-averaged pose and energy:
\begin{align}
\bar{E}(u) &= \frac{1}{w}\sum_{i=0}^{w-1} E_{u+i},
&
\bar{y}(u) &= \frac{1}{w}\sum_{i=0}^{w-1} \tilde{y}_{u+i}.
\end{align}
We then apply the same pairwise form:
\begin{equation}
\ell_{\mathrm{pair}}(t,s)
=
\Bigl[\kappa\cdot \mathrm{MPJPE}\!\bigl(\bar{y}(t),\bar{y}(s)\bigr)
-\bigl| \bar{E}(t)-\bar{E}(s)\bigr|\Bigr]_+.
\end{equation}
Detailed ablation results are reported in Appendix~\ref{appendix:abl}.

\section{Experimental Setup}
\subsection{Datasets and Evaluation Metrics}
\paragraph{Datasets.}
We conduct our empirical experiments on Human3.6M dataset (H36M)~\citep{h36m_pami}, MPI-INF-3DHP (3DHP)~\citep{3dhp} dataset and Human3.6M 3D WholeBody dataset (H3WB)~\citep{h3wb}.
H36M is the most widely used dataset for 3D human pose estimation~\citep{survey2, survey1}. 3DHP is a more challenging dataset than H36M because it contains fewer samples and includes both indoor and outdoor scenes, while H36M only contains indoor scenes. H3WB is a recent dataset for 3D whole-body pose estimation. H3WB extends H36M by providing whole-body keypoint annotations with detailed information about hands, face, and feet, making it suitable for evaluating fine-grained 3D pose estimation. Following common practice on 3DHP, which is typically evaluated using ground-truth 2D inputs, we also report results on H36M with ground-truth 2D keypoints for a fair comparison across datasets.
\paragraph{Evaluation Metrics.}
We follow common practice in 3D human pose estimation and report standard metrics, such as mean per-joint position error (MPJPE), procrustes-aligned MPJPE (P-MPJPE), percentage of correct keypoints (PCK), area under curve (AUC), and pelvis-aligned MPJPE (PA-MPJPE), according to the evaluation protocol of each dataset. 
These metrics remain the standard benchmarks to evaluate per-joint error. However, they do not measure structural plausibility, whether the predicted poses conform to anatomical constraints.

\subsection{Structural Consistency Metrics}
\label{subsec:structural_consistency}
To further evaluate structural consistency, we introduce two additional metrics.\\
\textbf{Limb Symmetry Error (LSE).}
LSE measures violation of the left–right symmetry by comparing the lengths of the corresponding limbs, such as the lower arms and thighs. For a limb pair $(\mathbf{l}_{i1}, \mathbf{l}_{i2}), (\mathbf{r}_{i1}, \mathbf{r}_{i2})$, LSE is defined as the normalized difference in length between the left and right counterparts:
\[
\text{LSE}_i = 100 \cdot \left| \frac{ \| \mathbf{l}_{i1} - \mathbf{l}_{i2} \| - \| \mathbf{r}_{i1} - \mathbf{r}_{i2} \| }{ ( \| \mathbf{l}_{i1} - \mathbf{l}_{i2} \| + \| \mathbf{r}_{i1} - \mathbf{r}_{i2} \| ) / 2 } \right|
\]

\paragraph{Body Segment Length Error (BSLE).}
BSLE measures deviations in the lengths of body segments, pair of adjacent joints, by comparing predicted poses and ground-truth poses. A special case of BSLE focuses on limbs, where symmetric differences are most pronounced, is referred to as \textbf{limb length error (LLE)}. For each segment $i$, with predicted adjacent keypoints $\mathbf{k}_{i1}, \mathbf{k}_{i2}$ and corresponding ground truth keypoints $\mathbf{t}_{i1}, \mathbf{t}_{i2}$, BSLE is defined as:
\[
\text{BSLE}_i = 100 \cdot \left| 1 - \frac{ \| \mathbf{k}_{i_2} - \mathbf{k}_{i_1} \|}{\| \mathbf{t}_{i_2} - \mathbf{t}_{i_1} \|} \right|
\]
We emphasize that these metrics are not used as training loss. Instead, our trainable loss-net is designed to learn structural consistency directly from data without requiring explicit priors such as fixed bone lengths or symmetry constraints. LSE and BSLE are used solely for evaluation purposes, providing complementary insights into whether predicted poses are anatomically plausible.

\subsection{Backbone models}
We employed a diverse set of pose estimation models as our pose-net backbones, ranging from general baselines to recent state-of-the-art methods. Specifically, we used SimpleBaseline~\citep{simplebaseline}, SemGCN~\citep{semgcn}, and VideoPose~\citep{videopose} for single-frame setting, and MixSTE~\citep{mixste}, P-STMO~\citep{pstmo}, PoseFormerV2~\citep{poseformerv2}, D3DP~\citep{shan2023diffusion}, KTPformer~\citep{peng2024ktpformer} and MotionAGFormer~\citep{Mehraban_2024_WACV} for multi-frame setting. These backbones cover a broad range of architectures, from commonly used designs to state-of-the-art models, allowing us to verify the effectiveness and robustness of \sealp~across different designs. Further implementation details are provided in the Appendix~\ref{details}.

\section{Experimental Results}
We provide a comprehensive set of 34 experimental results (single-frame models: 3 baselines × 3 datasets × 2 \sealp~variants; multi-frame models: 4 baselines × 2 variants on H36M and 3DHP), which consistently corroborate our findings and offer strong evidence for the robustness and generalizability of our approach.
\subsection{Comparison with Baselines}
\label{result:main}
\begin{table*}[t]
  \centering
  \begin{minipage}[t]{0.48\textwidth}
    \centering
    \scriptsize
    \setlength{\tabcolsep}{8pt}
    \caption{\textbf{Performances on Human3.6M}. \sealp~improves MPJPE and P-MPJPE across models. D3DP and KTPformer use $T{=}243,H{=}20,K{=}10,J_{Best}$.}
    \label{tab:h36m}
    \resizebox{\linewidth}{!}{
    \begin{tabular}{@{}lcc@{}}
      \toprule
      Method & MPJPE$\downarrow$ & P\hbox{-}MPJPE$\downarrow$ \\
      \midrule
      \multicolumn{3}{@{}l}{\emph{Single-frame models}} \\
      \midrule
      SimpleBaseline~\citep{simplebaseline}                 & 43.8 & 34.7 \\
      \quad + \sealp~(MLP)        & 42.5 & 33.9 \\
      \quad + \sealp~(Graph)      & \textbf{40.7} & \textbf{32.3} \\
      \midrule
      SemGCN~\citep{semgcn}                         & 47.0 & 37.9 \\
      \quad + \sealp~(MLP)        & 44.9 & 36.5 \\
      \quad + \sealp~(Graph)      & \textbf{43.4} & \textbf{35.7} \\
      \midrule
      VideoPose~\citep{videopose}                      & 41.6 & 32.4 \\
      \quad + \sealp~(MLP)        & \textbf{41.0} & 32.3 \\
      \quad + \sealp~(Graph)      & 41.2 & \textbf{32.1} \\
      \toprule
      Method & MPJPE$\downarrow$ & P\hbox{-}MPJPE$\downarrow$ \\
      \midrule
      \multicolumn{3}{@{}l}{\emph{Multi-frame models}} \\
      \midrule
      MixSTE ($T{=}243$)~\citep{mixste}             & 20.8 & 16.1 \\
      \quad + \sealp~(MLP)        & 20.6 & 15.8 \\
      \quad + \sealp~(Graph)      & \textbf{20.0} & \textbf{15.7} \\
      \midrule
      Poseformer V2 ($T{=}27$)~\citep{poseformerv2}  & 42.7 & 31.6 \\
      \quad + \sealp~(MLP)        & 41.5 & 31.2 \\
      \quad + \sealp~(Graph)      & \textbf{40.5} & \textbf{30.3} \\
      \midrule
      D3DP~\cite{shan2023diffusion}  & 20.4 &  15.4\\
      \quad + \sealp~(MLP)        &   18.1 & 13.9 \\
      \quad + \sealp~(Graph)      & \textbf{17.7} & \textbf{13.7} \\
      \midrule
      KTPformer~\cite{peng2024ktpformer}  & 18.9 & 14.3  \\
      \quad + \sealp~(MLP)        &  18.9  & 14.5 \\
      \quad + \sealp~(Graph)      & \textbf{18.3} & \textbf{13.9}  \\
      \bottomrule
    \end{tabular}
    }%
  \end{minipage}%
  \hfill
  \begin{minipage}[t]{0.48\textwidth}
    \centering
    \scriptsize
    \setlength{\tabcolsep}{2pt}
    \sisetup{detect-weight=true, detect-family=true, mode=text}
    \caption{\textbf{Performances on MPI-INF-3DHP}. \sealp~consistently reduces MPJPE and improves PCK and AUC. D3DP uses $T{=}243,H{=}20,K{=}20,J_{Best}$.}
    \label{tab:3dhp}
    \setlength{\tabcolsep}{4.45pt}
    \resizebox{\linewidth}{!}{%
    \begin{tabular}{
      @{}l
      S[table-format=2.1]
      S[table-format=2.1]
      S[table-format=2.1]
      @{}
    }
    \toprule
    {Method} & {MPJPE$\downarrow$} & {PCK$\uparrow$} & {AUC$\uparrow$} \\
    \midrule
    \multicolumn{4}{@{}l}{\emph{Single-frame models}} \\
    \midrule
    SimpleBaseline~\citep{simplebaseline}                 & 80.9 & 86.9 & 53.8 \\
    \quad + \sealp~(MLP)        & 71.8 & 89.3 & 58.7 \\
    \quad + \sealp~(Graph)      & \textbf{68.2} & \textbf{90.2} & \textbf{60.4} \\
    \midrule
    SemGCN~\citep{semgcn}                        & 74.5 & 89.5 & 56.4 \\
    \quad + \sealp~(MLP)        & 71.8 & 90.4 & 57.9 \\
    \quad + \sealp~(Graph)      & \textbf{62.9} & \textbf{92.7} & \textbf{61.7} \\
    \midrule
    VideoPose~\citep{videopose}                      & 66.4 & 90.8 & 60.5 \\
    \quad + \sealp~(MLP)        & 64.0 & 91.7 & 62.1 \\
    \quad + \sealp~(Graph)      & \textbf{62.2} & \textbf{91.8} & \textbf{63.1} \\
    \toprule
    {Method} & {MPJPE$\downarrow$} & {PCK$\uparrow$} & {AUC$\uparrow$} \\
    \midrule
    \multicolumn{4}{@{}l}{\emph{Multi-frame models}} \\
    
    \midrule
    P\hbox{-}STMO ($T{=}81$)~\citep{pstmo}      & 33.4 & 98.0 & 77.5\\
    \quad + \sealp~(MLP)         & 32.9 & 98.1 & 77.7 \\
    \quad + \sealp~(Graph)       & \textbf{32.5} & \textbf{98.2} & \textbf{78.0} \\
    \midrule
    Poseformer V2 ($T{=}27$)~\citep{poseformerv2}      & 29.6 & 97.0 & 77.8 \\
      \quad + \sealp~(MLP)        & 28.5 & 97.4 & 78.4 \\
      \quad + \sealp~(Graph)      & \textbf{27.8} & \textbf{97.5} & \textbf{79.0} \\
    \midrule
    D3DP~\cite{shan2023diffusion}  & 28.8 & 98.2  & 80.4\\
      \quad + \sealp~(MLP)        & 28.5 & 98.6  & 80.8 \\
      \quad + \sealp~(Graph)      & \textbf{27.8}  & \textbf{98.7} & \textbf{80.9} \\
    \midrule
    MotionAGFormer-L~\cite{Mehraban_2024_WACV}      & 16.2 & 98.2  & 85.3 \\
      \quad + \sealp~(MLP)        & \textbf{15.2} & \textbf{99.1} & \textbf{89.0} \\
      \quad + \sealp~(Graph)      & 15.3 & 99.0 & 88.9 \\
    \bottomrule
    \end{tabular}
    }%
  \end{minipage}
\end{table*}
\begin{table}[t]
\centering
\caption{\textbf{Performance on the Human3.6M WholeBody}. \sealp~reduces P-MPJPE across all body parts, resulting in more coherent predictions. $\dagger$ \footnotesize from H3WB's official benchmark. $\ddagger$ nose-aligned MPJPE for face and wrist-aligned MPJPE for hands. 
}
\label{tab:h3wb}
\small
\setlength{\tabcolsep}{2pt}
\resizebox{\columnwidth}{!}{
\begin{tabular}{@{}lcccc@{}}
\toprule
Method                     & Whole-body & Body      & Face/Aligned\textsuperscript{$\ddagger$} & Hand/Aligned\textsuperscript{$\ddagger$} \\ \midrule
Jointformer \textsuperscript{$\dagger$} & 
    88.3 &
    84.9 &
    66.5 / 17.8 &
    125.3 / 43.7 \\ 
3D-LFM~\small &
    64.1 &
    60.8 &
    56.6 / 10.4 &
    78.2 / 28.2 \\  
\midrule
SimpleBaseline~\citep{simplebaseline}              & 67.4          & 63.3          & 49.9 / 14.1  & 98.0 / 34.8          \\
+ \sealp~(MLP)        & \textbf{62.8}         & \textbf{61.1}          & \textbf{46.3} / 13.7  & \textbf{90.7} / 34.2          \\
+ \sealp~(Graph)      & 64.8 & 61.6                                    & 47.6 / \textbf{13.3}  & 94.0 / 34.5 \\ 
\midrule
VideoPose~\citep{videopose}                   & 61.5          & 57.4          & 48.8 / 11.9  & 84.1 / 30.3          \\
+ \sealp~(MLP)        & \textbf{58.6}          & \textbf{54.8}          & \textbf{45.0} / \textbf{11.5}   & \textbf{82.3} / 29.3          \\
+ \sealp~(Graph)    & 59.5          & 56.3          & 46.1 / \textbf{11.5}   & \textbf{82.3} / 28.7          \\ \bottomrule
\end{tabular}
}
\end{table}

\textbf{Single$\text{-}$Frame Settings.}
As shown in Table~\ref{tab:h36m} (Human3.6M) and Table~\ref{tab:3dhp} (MPI-INF-3DHP), \sealp~consistently improves performance across all baseline models. 
On Human3.6M, \sealp~reduces both MPJPE and P-MPJPE, with graph-based loss-net variants yielding the largest gains.
On MPI-INF-3DHP, which contains more diverse and In-The-Wild scenarios, the improvements are even more pronounced. \sealp~also improves whole-body pose estimation on the H3WB dataset, as shown in (Table~\ref{tab:h3wb}), demonstrating that our trainable loss function is also beneficial in complex whole-body settings. 
However, when the loss-net is graph-based, its bias toward local neighborhoods may underutilize global context and thus underperform a simpler MLP loss-net. We leave a deeper analysis of the balance between local and global reasoning for future work.

\textbf{Multi$\text{-}$Frame Settings.} 
Multi-frame models already perform strongly by leveraging temporal information, and recent architectures have largely reached a saturated regime in benchmark performance. In this work, we apply \sealp\ to a broad set of multi-frame backbones, including general transformer-style baselines (P-STMO, MixSTE, PoseFormerV2), a diffusion-based model (D3DP), and recent state-of-the-art methods (KTPFormer, MotionAGFormer). Despite the strength of these models, adding our loss-net consistently yields additional improvements across all backbones, as shown in Table~\ref{tab:h36m} and Table~\ref{tab:3dhp}.

\begin{figure}[t]
    \centering
    \includegraphics[width=1\linewidth]{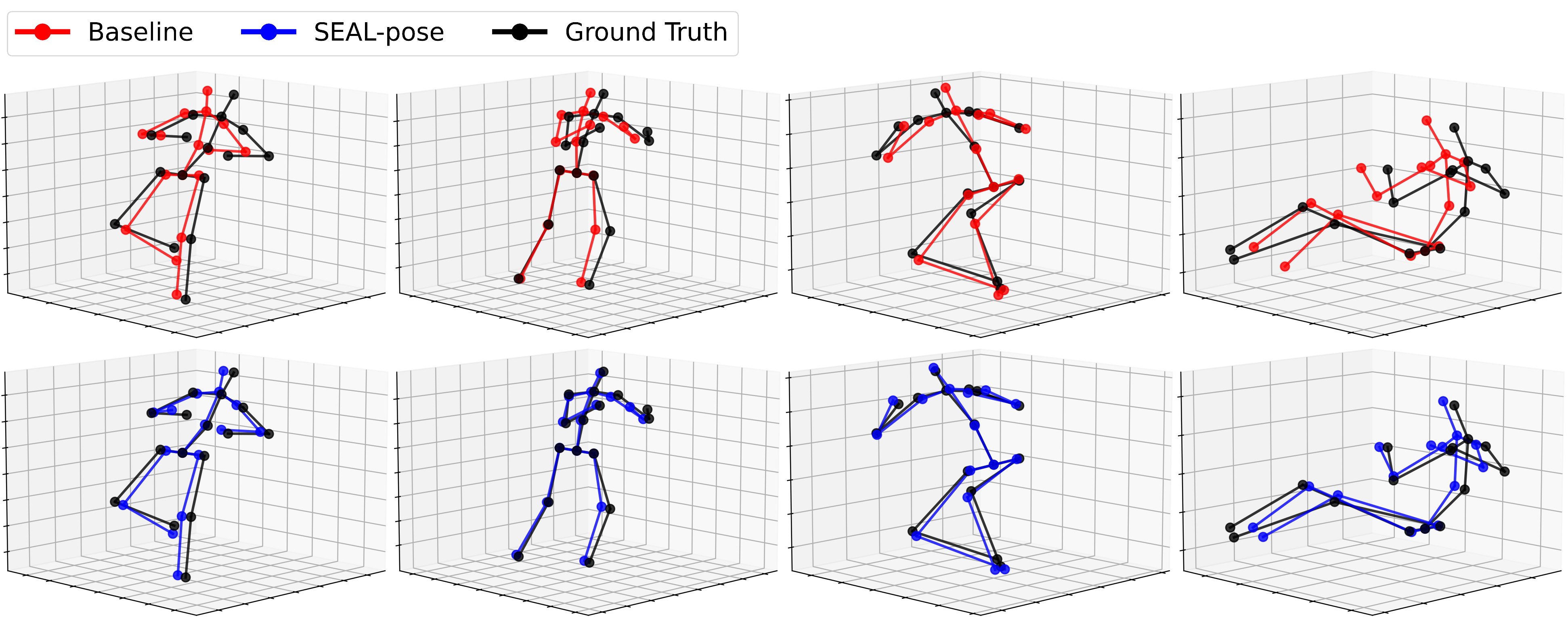}
    \caption{\textbf{Qualitative Comparison of Predicted Poses on H36M}. 
    Predictions from \sealp~(bottom, blue) demonstrate clear improvements over the baseline (top, red) by producing structures closer to the ground-truth human pose (black).
    }
    \label{fig:example-output}
\end{figure}

\subsection{Training stability and hyperparameter search}
\label{subsec:trian_stability}

The convergence trends in Fig.~\ref{fig:train-mpjpe} further support stable optimization.
We retain an explicit supervised regression loss in $\mathcal{L}_F$ as an optimization anchor, and use the energy term only as an auxiliary regularizer that shapes structural plausibility; this anchor stabilizes training and yields reliable convergence.

Building on this stability, rather than exhaustively searching the large hyperparameter space, we adopt a simple greedy tuning protocol (Appendix~\ref{appendix:greedy}), which is efficient in practice.
To further verify robustness, we additionally evaluate several nearby sub-optimal settings; Table~\ref{tab:hp_sweep} and Fig.~\ref{fig:train-energy} show consistent convergence behavior without instability.

\subsection{Structural Consistency Evaluation}
\label{result:structure}
We evaluated structural consistency by examining the LSE, LLE, and BSLE metrics on the H36M, 3DHP, and H3WB datasets. For comparison, we also included a setting with explicit bone length constraints as a loss term, to directly contrast \sealp~with manually designed constraint-based approaches.

\begin{table}[t]
\centering
\caption{\textbf{Structural Consistency Evaluation Across Datasets}.
\sealp~reduces structural error metrics such as LSE, LLE, and BSLE (see~\sref{subsec:structural_consistency} for the definitions), improving plausibility.}
\label{tab:metric}
\small
\setlength{\tabcolsep}{2pt}
\resizebox{\linewidth}{!}{
\begin{tabular}{@{}llcccc@{}}
\toprule
Dataset & Metric & MPJPE$\downarrow$ & LSE$\downarrow$ & LLE$\downarrow$ & BSLE$\downarrow$ \\
\midrule
H36M & Ground Truth &  & 0.00 & 0.00 & 0.00 \\
     & SimpleBaseline~\citep{simplebaseline} & 43.8 & 4.85 & 5.09 & 6.12 \\
     & + Constraint~\citep{lbone}            & 41.5  & 4.35 & 4.54 & 6.17 \\
     & + \sealp~                      & \textbf{40.7}  & \textbf{3.68} & \textbf{3.94} & \textbf{5.49} \\
\midrule
3DHP & Ground Truth &  & 1.21 & 0.00 & 0.00 \\
     & SimpleBaseline~\citep{simplebaseline} & 80.9 & 10.14 & 11.60 & 8.13 \\
     & + Constraint~\citep{lbone}            & 75.0 & 7.80  & 10.02 & 7.87 \\
     & + \sealp~                    & \textbf{68.2} & \textbf{6.22} & \textbf{6.02} & \textbf{5.93} \\
\midrule
H3WB & Ground Truth &  & 4.42 & 0.00 & 0.00 \\
     & SimpleBaseline~\citep{simplebaseline} & 67.4 & 6.60 & 6.56 & \textbf{6.22} \\
     & + Constraint~\citep{lbone}            & 65.1 & 6.88 & 6.82 & 6.66 \\
     & + \sealp~                      & \textbf{62.8} & \textbf{6.55} & \textbf{6.13} & 6.73 \\
\bottomrule
\end{tabular}
}
\end{table}

In result, \sealp~consistently showed lower error values across all three structural metrics on H36M and 3DHP datasets, as detailed in Table~\ref{tab:metric}. These results indicate that \sealp~effectively captures structured dependencies in human poses, leading to more anatomically plausible and consistent 3D pose predictions. 
For a more detailed examination, we grouped samples into bins with comparable P-MPJPE and analyzed our proposed structural consistency metrics (LSE, BSLE, LLE) within each bin. Even when comparing predictions with similar P-MPJPE, \sealp~consistently yielded lower values on these metrics than the baseline, as shown in Fig.~\ref{fig:struct-err-3dhp}. 
These intra-bin comparisons reveal a limitation of standard training objective: by optimizing primarily for pointwise coordinate error (e.g. MSE, MPJPE), they are relatively insensitive to violations of structural plausibility (e.g., symmetry, limb proportionality, kinematic consistency), leading to higher LSE/BSLE/LLE within similar P-MPJPE ranges.
Notably, LSE is defined without reference to ground truth, yet the results on 3DHP highlight that \sealp~more effectively internalizes structural constraints. On Human3.6M, absolute errors are already small, so the differences are less pronounced, but \sealp~still demonstrates an advantage over the baseline.

However, \sealp~showed mixed results on the H3WB dataset, 
This is likely due to the dataset’s noisy labeling, which is shown from the relatively high LSE of the ground truth poses. 
Moreover, since H3WB provides annotations for a very large number of keypoints including body, face, hands, and feet, capturing coherent structural dependencies across all regions is inherently more challenging, making further improvements less straightforward. 
While \sealp~did not significantly outperform the baseline in this setting, it still achieved better results compared to directly injecting explicit structural constraints, suggesting that a trainable loss function provides a more flexible and generalizable way of enforcing plausibility in whole-body pose estimation. Overall, the improved structural consistency metrics highlight that loss-net's ability to capture structures in human poses helps the pose-net to predict more anatomically consistent and plausible 3D human poses.

\subsection{Ablation Studies}

\begin{figure}
    \centering
    \includegraphics[width=1\linewidth]{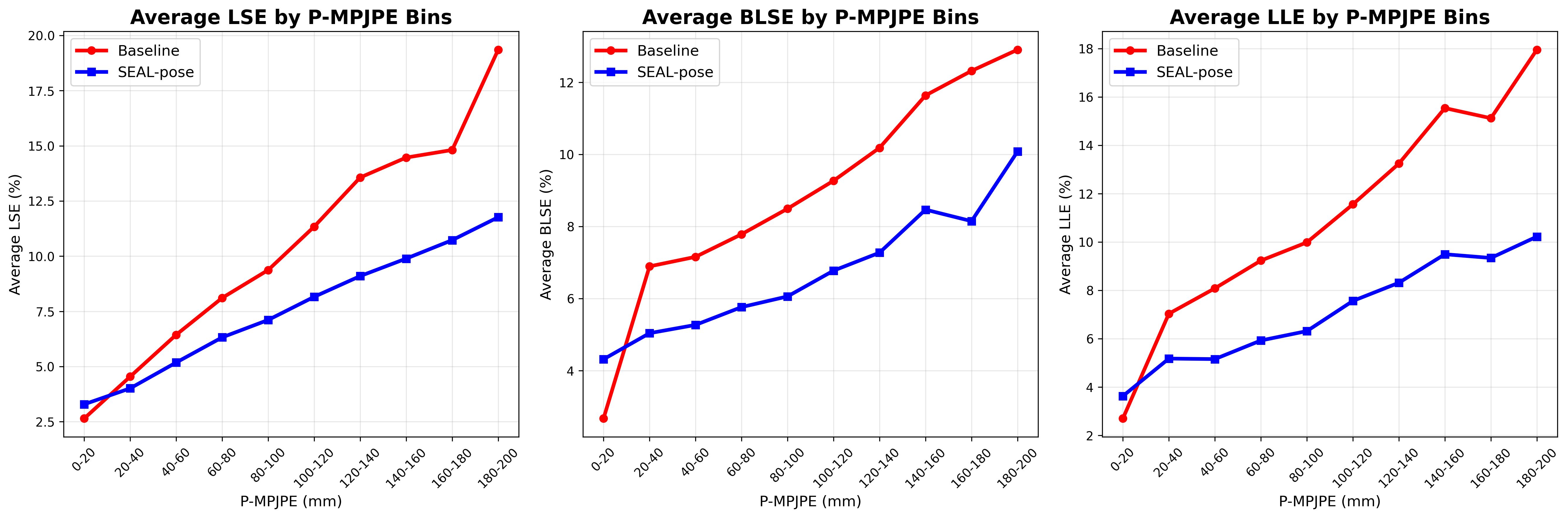}
    \caption{\textbf{Comparison of structural consistency in MPI-INF-3DHP.} Average structural inconsistency measures (from left to right: LSE, BSLE, LLE)   are displayed for predictions of baselines (\textcolor{red}{red}) and \sealp~(\textcolor{blue}{blue}) binned by P-MPJPE. \sealp~consistently achieves lower structural errors even under similar P-MPJPEs.}
    \label{fig:struct-err-3dhp}
\end{figure}


\paragraph{Analysis of the Energy-Guided Pose-Net.}We conduct gradient-based inference (GBI) on the output of the pose-net using the trained loss-net to verify its ability to capture plausible human pose structures. 
Starting from the pose-net prediction, GBI iteratively refines the pose by following gradient signals from the loss-net, which are expected to lower the assigned energy (Appendix~\ref{appendix:gbi}). Each iteration corresponds to a single training-style update step, and the curves—averaged over all test examples—show gradual improvement rather than drastic one-step changes. Since these metrics directly reflect structural plausibility, the consistent reduction indicates that the loss-net effectively captures human pose structure and provides meaningful gradient signals. The effect is more pronounced on the challenging 3DHP dataset, but the same trend is also observed on H36M, as shown in Fig.~\ref{fig:gbi-h36m} in Appendix~\ref{appendix:gbi}, confirming the consistency of the results.
\begin{figure}
    \centering
    \includegraphics[width=1\linewidth]{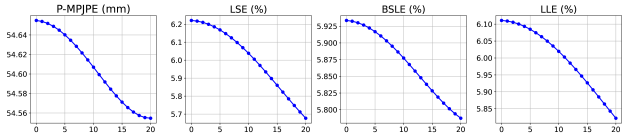}
    \caption{\textbf{Gradient-Based Inference results on MPI-INF-3DHP}. P-MPJPE, LSE, LLE, and BSLE all decrease steadily over iterations, indicating that the loss-net effectively captures structural plausibility and provides meaningful corrective feedback to the pose-net.}
    \label{fig:gbi-3dhp}
\end{figure}
\paragraph{Analysis of the Learned Energy.} We further examine whether the loss-net energy aligns with structural consistency metrics. The graph-based loss-net energy exhibits a positive correlation with the proposed structural metrics (LSE/BSLE/LLE), implying that higher energy corresponds to less structurally plausible poses. For instance, for LSE we observe Kendall's $\tau \approx 0.29$, which corresponds to roughly $65\%$ correct pairwise orderings. For the other metrics, we obtain $59.1\%$ for BSLE and $60.6\%$ for LLE.
 This suggests that the energy correlates with global structural plausibility, especially symmetry consistency.

\paragraph{Analysis of Graph-based Architecture.}We compared graph-based and MLP-based loss-nets. Overall results show that the graph-based loss-net consistently provides better structural guidance, leading to lower per-joint errors and improved structural plausibility, as shown in Sections~\ref{result:main} and~\ref{result:structure}. This trend is also reflected in the pairwise ordering analysis: the graph-based loss-net achieves $\sim$10\% higher correct pairwise orderings than the MLP-based loss-net across our structural metrics. This suggests that the graph-based loss-net provides more suitable training signals for learning structure-aware guidance.



\paragraph{Cross-Dataset Evaluation for Generalization and Robustness}To test whether the loss-net overfits the training data, we run cross-dataset experiments between Human3.6M (H36M) and MPI-INF-3DHP (3DHP), training on one dataset and evaluating on the other. If the loss-net were merely memorizing domain-specific patterns, adding \sealp~would be expected to worsen cross-dataset performance relative to the baseline, which would provide direct evidence of overfitting.
These results highlight two key points. First, the cross-dataset experiments indicate that limited domain generalization and sensitivity to input formatting are common issues in 3D HPE models, as both the baseline and \sealp~suffer from dataset shift.
Second, regarding the original concern about the overfitting of the loss-net, we actually observe the opposite behavior as the gain of \sealp~over the baseline increases under the domain shift (Fig.~\ref{fig:cross_dataset_hists}, Table~\ref{tab:cross_dataset_poseformerv2_mpjpe_pmpjpe}).
This suggests that the loss-net does not overfit in a way that harms the pose-net’s cross-dataset performance. 
Finally, we complement the cross-dataset evaluation with qualitative results on In-The-Wild images to further assess robustness. As shown in Appendix~\ref{appendix:inthewild} (Fig.~\ref{fig:inthewild},~\ref{fig:inthewild2}), \sealp~yields more structurally consistent poses than the baseline(PoseFormerV2, KTPFormer) on the In-The-Wild images.
}

\begin{table}[ht]
  \centering
  \caption{Cross-dataset evaluation between Human3.6M (H36M) and MPI-INF-3DHP (3DHP) using PoseFormerV2. \sealp~does not introduce additional degradation beyond the baseline under dataset shift, suggesting no harmful loss-net overfitting.}
  \label{tab:cross_dataset_poseformerv2_mpjpe_pmpjpe}

  \footnotesize 
  \setlength{\tabcolsep}{3pt}

  \resizebox{\linewidth}{!}{%
    \begin{tabular}{lcc|cc}
      \toprule
      \multirow{2}{*}{Train $\rightarrow$ Test}
        & \multicolumn{2}{c|}{MPJPE$\downarrow$}
        & \multicolumn{2}{c}{P-MPJPE$\downarrow$}\\
      \cmidrule(lr){2-3}\cmidrule(lr){4-5}
        & Baseline & \sealp~
        & Baseline & \sealp~ \\
      \midrule
      H36M $\rightarrow$ 3DHP & 111.4 & \textbf{97.3} & 83.2 & \textbf{70.3} \\
      3DHP $\rightarrow$ H36M & 157.0 & \textbf{151.9} & 109.7 & \textbf{107.9} \\
      \bottomrule
    \end{tabular}%
  }
\end{table}


\section{Conclusion}

In this paper, we propose \sealp, a novel framework that introduces a trainable loss function for 3D human pose estimation. Unlike prior approaches relying on explicit priors or architecture-specific constraints, our proposed loss-net learns structural dependencies directly from data, and can be seamlessly integrated with a wide range of backbone pose-nets. 
We design the loss-net as a graph-based model, representing joints as nodes and bones as edges, which enables principled learning of both local (bone lengths, adjacency) and global (symmetry, long-range relations) dependencies in human pose.
Our experiments on Human3.6M, MPI-INF-3DHP, and H3WB demonstrate that \sealp~not only reduces per-joint pose estimation errors, but also improves structural plausibility, confirmed by our proposed LSE and BSLE metrics, showing that the graph-based loss-net provides stronger structural guidance.
Overall, \sealp~highlights the promise of trainable loss functions as a general paradigm for structured prediction tasks with complex output dependencies.

\clearpage
\section{Impact Statements}
This paper proposes \sealp, a trainable loss-function framework for 3D human pose estimation that learns structural dependencies among joints directly from data. By training a pose-net using a loss-net that evaluates structural plausibility, \sealp reduces per-joint errors while improving the overall structural validity of predicted poses. This can enable more reliable pose estimation for downstream applications such as motion analysis, human–computer interaction, sports analytics, rehabilitation/healthcare, and animation.

As with human-centric perception models in general, the proposed approach could be misused for privacy-invasive purposes such as surveillance or tracking, and dataset biases may lead to performance disparities under certain environments or conditions. Future work should strengthen robustness evaluation across diverse datasets and in-the-wild settings, and deployment should follow consent-based and privacy-preserving principles.

\bibliography{example_paper}
\bibliographystyle{icml2026}
\clearpage

\appendix
\section{Appendix}
\subsection{Detailed \sealp~}
\label{SEAL-pose-detail}
In our framework, the pose-net $F_\phi(x)$ is optimized to minimize a weighted sum of the mean squared error (MSE) loss and the output of the loss-net (energy) $E_\theta(x, \tilde{y})$. Specifically, the pose-net parameters $\phi$ are updated in the following manner:
\begin{align}
\phi_t \leftarrow \phi_{t-1} - \eta_\phi \nabla_\phi \frac{1}{|B_t|} \sum_{(x,y) \in B_t} L_F(\phi; \theta),
\label{eqn:1}
\end{align}
where $B_t$ is the mini-batch of training samples at iteration $t$, $\eta_\phi$ is the learning rate for the pose-net, and $L_F(\phi; \theta)$ is the combined loss function. The combined loss function is defined as:
\begin{align}
L_F(x_i, y_i; \theta) = \sum_{j=1}^M \text{MSE}(y_j, F_\phi(x)_j) + \alpha E_\theta(x, F_\phi(x)),
\label{eqn:2}
\end{align}
where $M$ refers to the total number of joints in the pose estimation dataset and $x$ represents the input data, specifically the 2D joint coordinates. The variable $y_j$ denotes the ground-truth 3D joint coordinates, while $F_\phi(x)_j = \tilde{y}_j$ represents the predicted 3D joint coordinates from the pose-net.
The energy term $E_\theta(x, F_\phi(x))$ is computed by the loss-net and implicitly evaluates the structural dependencies between joints.
Finally, $\alpha$ is a hyperparameter controlling the balance between the MSE loss and the energy term. 

The loss-net is dynamically trained to adapt to the pose-net's predictions by minimizing the loss $L_E$:
\begin{align}
\theta_t \leftarrow \theta_{t-1} - \eta_\theta \nabla_\theta \frac{1}{|B_t|} \sum_{(x,y) \in B_t} L_E(x, y, F_{\phi_{t-1}}(x); \theta).
\label{eqn:3}
\end{align}

We employ two types of loss for $L_E$: margin-based loss and a simplified form of noise contrastive estimation (NCE) ranking loss~\citep{nce}, both suggested in~\cite{seal}.

The margin-based loss enforces the loss-net to decrease the energy $E_\theta(x, y)$ of the ground truth label $y$ and increase the energy $E_\theta(x, \tilde{y})$ of the pose-net's incorrect prediction $\tilde{y}$, such that the difference between the two energies is sufficiently large to exceed the margin. The margin-based loss is defined as:
{\small
\begin{align}
L_E^{\text{margin}}(x_i, y_i, \tilde{y}_i; \theta) = \max_{\tilde{y}} \left[ \Delta(y, \tilde{y}) - E_\theta(x, \tilde{y}) + E_\theta(x, y) \right]_+,
\label{eqn:4}
\end{align}}where $\Delta(y, \tilde{y})$ denotes a task-specific margin function, MPJPE in our implementation.

Similarly, the NCE ranking loss minimizes the energy of the ground truth label $y$ while increasing the energy of the pose-net's prediction $\tilde{y}$, treating the pose-net's predictions as negative samples. The NCE ranking loss is defined as:
{\small \begin{align}
L_E^{\text{NCE}}(x_i, y_i, \tilde{y}_i; \theta) = - \log \frac{\exp(-E_\theta(x, y))}{\exp(-E_\theta(x, y)) + \exp(-E_\theta(x, \tilde{y}))}.
\label{eqn:5}
\end{align}}

\subsection{Implementation Details}
\label{details}
\subsubsection{Pose-Net}
We have modified the input and output layers of pose-net models to align with the dimensions of each dataset.
In single-frame settings, we used separate Adam optimizers~\citep{adam} without learning rate decay for the loss-net and pose-net and trained models with a batch size of 1024 for 50 epochs on H36M and 3DHP, and a batch size of 64 for 200 epochs on H3WB. For multi-frame backbones, we adopt the training configurations from the original works—i.e., the same training schedules, data augmentations, and hyperparameters as in their papers/official implementations. In our experiments, we do not retune backbone-specific settings; instead, we only vary three hyperparameters introduced by our method: the pose-net learning rate ($lr_p$), the loss-net learning rate ($lr_l$), and the energy weight ($\alpha$).

\subsubsection{Loss-Net}
\label{appendix:graphormer}
Graphormer injects structure into self-attention by adding shortest path distance (SPD)--based spatial bias \(b_{\phi(v_i,v_j)}\) and an edge-path bias \(c_{ij}\) (edges along a shortest path) to the attention logits:
\[
A_{ij}
= \frac{(h_i W_Q)(h_j W_K)^\top}{\sqrt{d}}
  + b_{\phi(v_i,v_j)} + c_{ij}.
\]
Separately, degree-based centrality is encoded by adding learnable embeddings to node inputs (not as an attention bias)~\citep{ying2021transformers}.
Because human skeletal graphs are small and regular compared with molecular or social networks, we adopt Graphormer’s core idea while simplifying it for human pose estimation. Concretely, we (i) remove node-level degree–centrality embeddings and (ii) do not define categorical edge types.
On small, skeletal graphs, such handcrafted encodings can act like noise and distract attention, so we eliminate them and let the model infer structure directly from data while retaining only the shortest-path biases.
This minimalist biasing reduces spurious inductive signals and helps the pose-net produce outputs with stronger structural consistency.
Beyond encoding local and global dependencies, Graphormer also introduces a global, CLS-like virtual node \(v_{\text{cls}}\) that aggregates whole-graph information~\citep{ying2021transformers}. We adopt this component in the loss-net: a virtual node \(v_{\text{cls}}\) attends to all joints and produces a compact summary signal of skeletal structure. This design summarizes pose-level structure and guides the pose-net toward structurally coherent outputs.

We design the graph-based loss-net, following the Graphormer foundation, with model width \(d{=}256\), \(H{=}8\) attention heads, depth \(=6\) blocks; the same graph bias is shared across layers. For inputs, we represent each node by concatenating the keypoint’s 2D coordinates with its predicted 3D coordinates and encoding the joint identity via a one-hot vector; the resulting feature is linearly projected to dimension \(d{=}32\), and a learnable CLS token is prepended. The head is an MLP that outputs a scalar energy, and we train the loss-net with either a margin-based objective (e.g., an MPJPE margin) or NCE. 

For the MLP loss-net, we adjusted the SimpleBaseline architecture by modifying the dimensions and depth of the hidden layers. Specifically, we set the hidden size to 2048 with 2 residual block stages and omitted batch normalization and dropout layers.

\subsection{Gradient-Based Inference}
\label{appendix:gbi}
\begin{figure}[ht]
    \centering
    \includegraphics[width=0.8\linewidth]{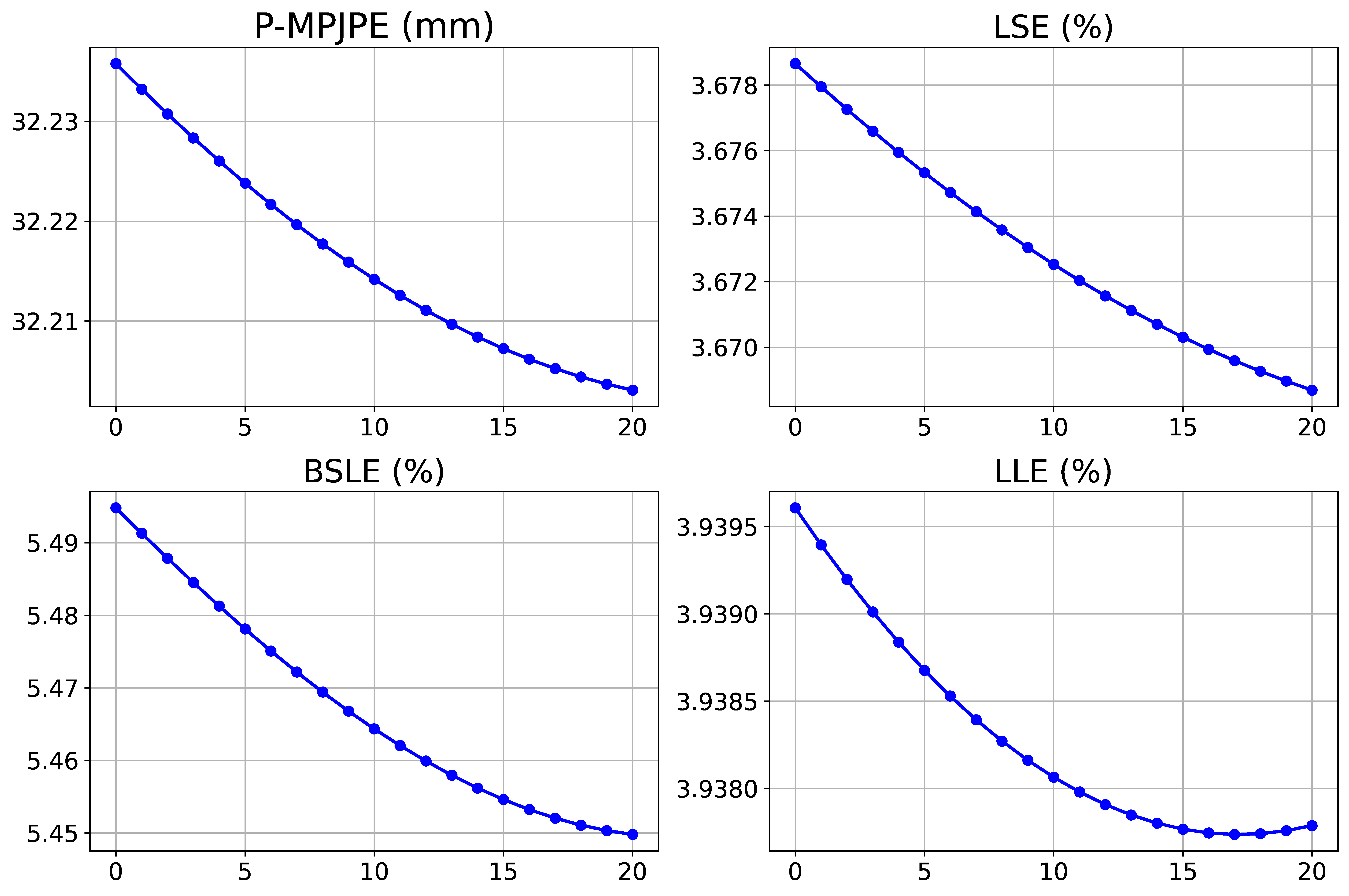}
    \caption{\textbf{Gradient-Based Inference results on Human3.6M}. P-MPJPE, LSE, LLE, and BSLE all decrease steadily over iterations, indicating that the loss-net effectively captures structural plausibility and provides meaningful corrective feedback to the pose-net.}
    \label{fig:gbi-h36m}
\end{figure}

We implement a gradient-based inference (GBI) method with trained loss-net and pose-net, to examine whether the loss-net effectively captures structural dependencies in human poses. 
GBI is a method that leverages gradients to iteratively refine the outputs~\citep{DBLP:journals/corr/GoodfellowSS14, Mordvintsev2015InceptionismGD,DBLP:journals/corr/GatysEB15a, DBLP:conf/nips/GatysEB15, spen} or parameters~\citep{gbi} of neural networks, and we adopt the former approach. Specifically, we iteratively update the predictions of pose-net along the gradient provided by the loss-net, with the objective of decreasing the energy. This procedure provides a direct way to evaluate whether the learned energy function captures human pose structure. If the loss-net has successfully learned structural dependencies, then following its gradient should progressively refine the predictions toward more plausible poses. The results on the H36M dataset are shown in Fig.~\ref{fig:gbi-h36m}, where all metrics steadily decrease over iterations, consistent with the trends observed for 3DHP in Fig.~\ref{fig:gbi-3dhp}.
\subsection{Result of Negative Sampling Strategies}
\label{appendix:abl}

In diffusion-based models, selecting the lowest-MPJPE sample as a hard negative yields the largest gain for multi-frame models, reducing MPJPE by $2.7$mm on H36M. In the single-frame setting, we generate synthetic negatives by perturbing 2D keypoints and forwarding them through the same deterministic pose-net. On 3DHP (MPJPE), the proposed negative--negative pairwise regularizer (Sec.~\ref{sec:synthetic-negatives}) provides additional gains of $2.6$mm for SimpleBaseline and $0.6$mm for SemGCN (Table~\ref{tab:sealp_3dhp_single_mpjpe}), while VideoPose shows no noticeable change. Finally, in the multi-frame setting, we further introduce a neighboring-window negative term that contrasts hypotheses across nearby frames; this yields a modest but consistent improvement over the strong MixSTE baseline (Table~\ref{tab:pair}), suggesting that cross-window negative contrasts can provide a complementary training signal to the structural guidance of the loss-net.

\begin{table}[ht]
\centering
\caption{\sealp~on 3DHP (single-frame models).
In parentheses: $\Delta$ MPJPE vs.\ each backbone baseline.}
\label{tab:sealp_3dhp_single_mpjpe}
\renewcommand{\arraystretch}{1.15}
\newcommand{\diff}[1]{{\scriptsize(#1)}}

\resizebox{\columnwidth}{!}{
\begin{tabular}{lccc}
\toprule
\textbf{Setting} & \textbf{SimpleBaseline} & \textbf{SemGCN} & \textbf{VideoPose} \\
\midrule
\sealp~(Graph)
& 68.2 \diff{-12.7}
& 62.9 \diff{-11.6}
& 62.2 \diff{-4.2} \\
\sealp~(Graph + pair-loss)
& 65.6 \diff{-15.3}
& 62.3 \diff{-12.2}
& 64.1 \diff{-2.3} \\
\bottomrule
\end{tabular}
}
\end{table}

\begin{table}[ht]
\centering
\caption{\sealp~on H36M with fixed $\Lambda$ and $K$ columns.
In parentheses: $\Delta$ vs.\ MixSTE ($T{=}243$); negative is better.}
\renewcommand{\arraystretch}{1.15}
\label{tab:pair}
\newcommand{\diff}[1]{{\scriptsize(#1)}}

\resizebox{\columnwidth}{!}{
\begin{tabular}{lcccccc}
\toprule
\textbf{Setting} & $\boldsymbol{\Lambda}$ & $\boldsymbol{K}$ & \textbf{$T$} &
\textbf{MPJPE$\downarrow$} & \textbf{P-MPJPE$\downarrow$} \\
\midrule
MixSTE (baseline) & --- & --- & 243 & 20.8 & 16.1  \\
\addlinespace[2pt]
\sealp~(Graph margin) & $10^{-3}$ & --- & 243
& 20.3 \diff{-0.5} & 15.8 \diff{-0.3}  \\
\sealp~(Graph margin + pair-loss) & $10^{-3}$ & 3 & 243
& \textbf{20.0} \diff{-0.8} & \textbf{15.7} \diff{-0.4}  \\
\bottomrule
\end{tabular}
}
\end{table}
\subsection{Simple greedy tuning protocol}
\label{appendix:greedy}
In practice, we adopt a simple greedy tuning protocol:
\begin{enumerate}
  \item We first tune the pose-net learning rate $lr_p$ (pose-net w/o loss-net),
  \item Given fixed $lr_p$, we explore the loss-net learning rate $lr_l$ and energy weight $\alpha$ then fix the best $lr_l$.
  \item Given fixed $lr_p$, $lr_l$, we finally tune the energy weight $\alpha$.
\end{enumerate}
When we apply the above greedy search,
we did not observe ``any'' training instability. The following With a concrete example on 3DHP with VideoPose~\citep{videopose}  shows the specific instance of greedy search over three hyperparameters $(lr_p, lr_l, \alpha)$. Please refer to Table~\ref{tab:hp_sweep}.
The best configuration we found was:
\begin{itemize}
  \item pose-net lr $(lr_p)$ = 1e-4
  \item loss-net lr $(lr_l)$ = 1e-4
  \item energy weight $(\alpha)$ = 5e-3
\end{itemize}
This setting yields a MPJPE of 62.18 ($<$ 66.4 of baseline).
Many nearby settings achieve very similar performance, within about 0.7 mm of the best score.
Across the broader grid of learning rate and energy weight,
the resulting MPJPE varies smoothly from roughly 62 mm up to approximately 67 mm as we move away from this well-tuned region, rather than collapsing or diverging. Regarding the specific question on the sensitivity to energy weight: for fixed learning rates $lr_p = lr_l = 1e{-4}$,
sweeping $\alpha$ over two orders of magnitude (from 1e-4 to 1e-2) changes MPJPE from 62.18 mm to at most 63.82 mm.
We find that energy weight $\alpha$ in the range of roughly 2e-3--5e-3 gives near-optimal performance, and values outside this band shift model smoothly rather than catastrophically. Similar patterns hold for other backbones and datasets.
\begin{figure}[ht]
    \centering
    \includegraphics[width=1\linewidth]{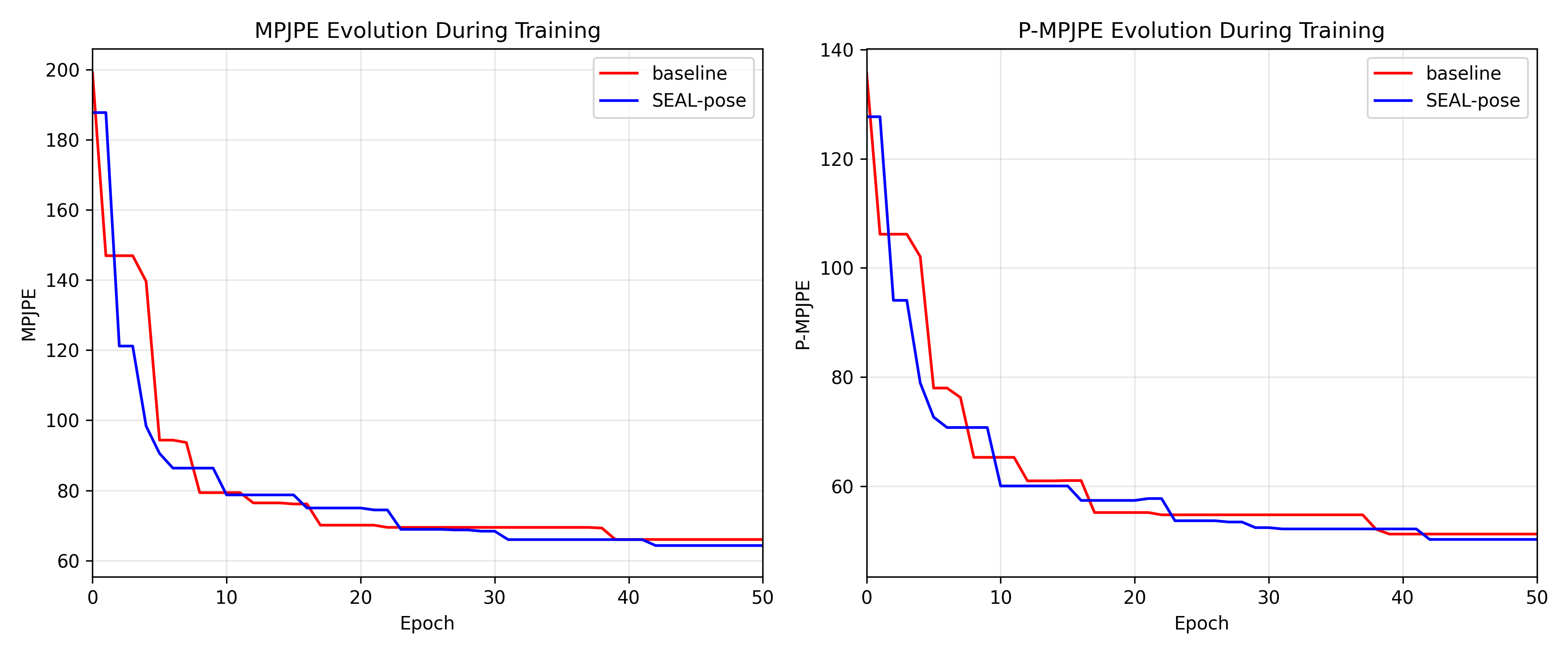}
    \caption{\textbf{Convergence comparison between the baseline and SEAL-pose during training.} Left: MPJPE, Right: P-MPJPE across epochs. Compared to the baseline (red), SEAL-pose (blue) exhibits similarly stable training dynamics while converging slightly faster, reaching comparable or lower error levels earlier.}
    \label{fig:train-mpjpe}
\end{figure}

\begin{figure}[ht]
    \centering
    \includegraphics[width=1\linewidth]{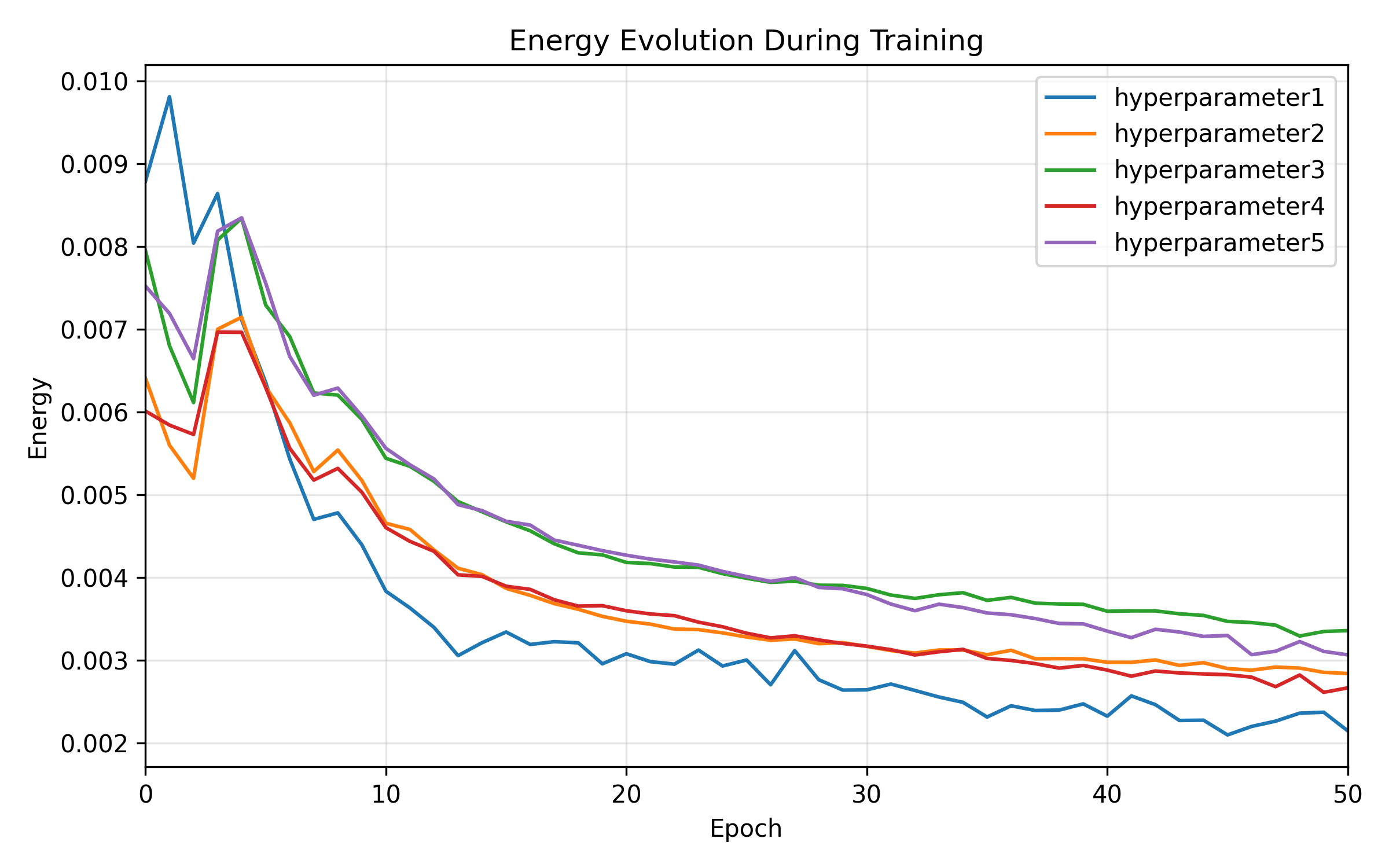}\caption{\textbf{Energy convergence across hyperparameter settings.} Across hyperparameters, the energy score consistently converges to a stable regime, suggesting that training remains stable and proceeds reliably after reaching a steady sub-optimal solution.}
    \label{fig:train-energy}
\end{figure}

\begin{table}[t]
\centering
\small
\setlength{\tabcolsep}{6pt}
\begin{tabular}{cccc}
\toprule
\textbf{pose-net lr} & \textbf{loss-net lr} & \textbf{energy weight} & \textbf{MPJPE} \\
\midrule
0.00010 & 0.00010 & 0.005   & 62.178 \\
0.00010 & 0.00010 & 0.002   & 62.308 \\
0.00005 & 0.00003 & 0.004   & 62.476 \\
0.00005 & 0.00003 & 0.005   & 62.505 \\
0.00010 & 0.00003 & 0.004   & 62.602 \\
0.00010 & 0.00003 & 0.005   & 62.719 \\
0.00010 & 0.00010 & 0.001   & 62.820 \\
0.00010 & 0.00005 & 0.001   & 62.841 \\
0.00010 & 0.00004 & 0.003   & 62.892 \\
0.00010 & 0.00005 & 0.005   & 62.919 \\
0.00010 & 0.00010 & 0.010   & 62.948 \\
0.00010 & 0.00004 & 0.005   & 62.955 \\
0.00010 & 0.00004 & 0.004   & 62.998 \\
0.00010 & 0.00010 & 0.00050 & 63.072 \\
0.00010 & 0.00005 & 0.002   & 63.079 \\
0.00010 & 0.00010 & 0.003   & 63.085 \\
0.00010 & 0.00002 & 0.005   & 63.236 \\
0.00010 & 0.00002 & 0.001   & 63.276 \\
0.00005 & 0.00010 & 0.005   & 63.356 \\
0.00005 & 0.00010 & 0.004   & 63.365 \\
0.00010 & 0.00003 & 0.001   & 63.381 \\
0.00010 & 0.00003 & 0.003   & 63.479 \\
0.00010 & 0.00001 & 0.005   & 63.520 \\
0.00010 & 0.00002 & 0.004   & 63.554 \\
0.00005 & 0.00050 & 0.004   & 63.561 \\
0.00010 & 0.00010 & 0.004   & 63.570 \\
0.00010 & 0.00003 & 0.002   & 63.580 \\
0.00020 & 0.00003 & 0.004   & 63.693 \\
0.00010 & 0.00005 & 0.004   & 63.782 \\
0.00050 & 0.00010 & 0.001   & 63.789 \\
0.00010 & 0.00010 & 0.00010 & 63.819 \\
0.00020 & 0.00005 & 0.001   & 63.895 \\
0.00005 & 0.00003 & 0.00050 & 64.006 \\
0.00005 & 0.00005 & 0.001   & 64.045 \\
0.00010 & 0.00004 & 0.006   & 64.092 \\
0.00010 & 0.00001 & 0.001   & 64.159 \\
0.00005 & 0.00003 & 0.00010 & 64.495 \\
0.00050 & 0.00003 & 0.004   & 64.555 \\
0.00005 & 0.00020 & 0.004   & 64.599 \\
0.00005 & 0.00003 & 0.001   & 64.639 \\
0.00005 & 0.00003 & 0.00005 & 64.751 \\
0.00005 & 0.00003 & 0.00020 & 64.948 \\
0.00005 & 0.00003 & 0.002   & 65.091 \\
0.00020 & 0.00003 & 0.005   & 65.358 \\
0.00050 & 0.00010 & 0.004   & 65.571 \\
0.00050 & 0.00003 & 0.005   & 66.290 \\
0.00050 & 0.00010 & 0.005   & 67.015 \\
0.00002 & 0.00003 & 0.004   & 67.233 \\
\bottomrule
\end{tabular}
\caption{\textbf{Hyperparameter sweep over learning rates(pose-net/loss-net) and energy weight.}}
\label{tab:hp_sweep}
\end{table}

\subsection{In-The-Wild Images visualization}
\label{appendix:inthewild}
\begin{figure*}[t]
    \centering
    \includegraphics[width=0.95\textwidth]{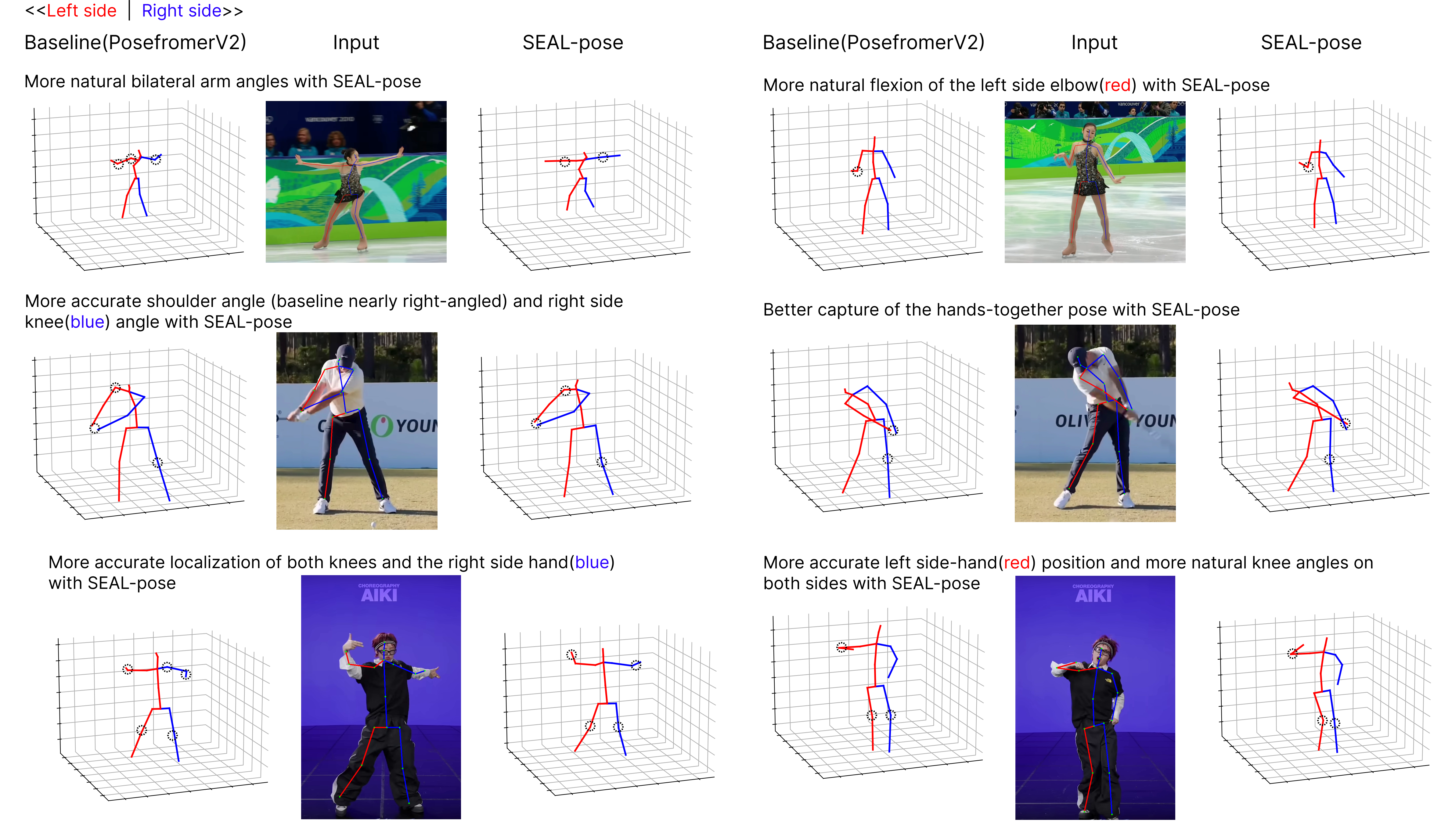}
    \caption{\textbf{Qualitative Comparisons of \sealp~with PoseFormerV2~\citep{poseformerv2} on In-The-Wild images.} Provides visualizations on unseen In-The-Wild images, illustrating that our method produces structurally plausible poses in out-of-distribution settings.}
    \label{fig:inthewild}
\end{figure*}

\begin{figure*}[t]
    \centering
    \includegraphics[width=0.95\textwidth]{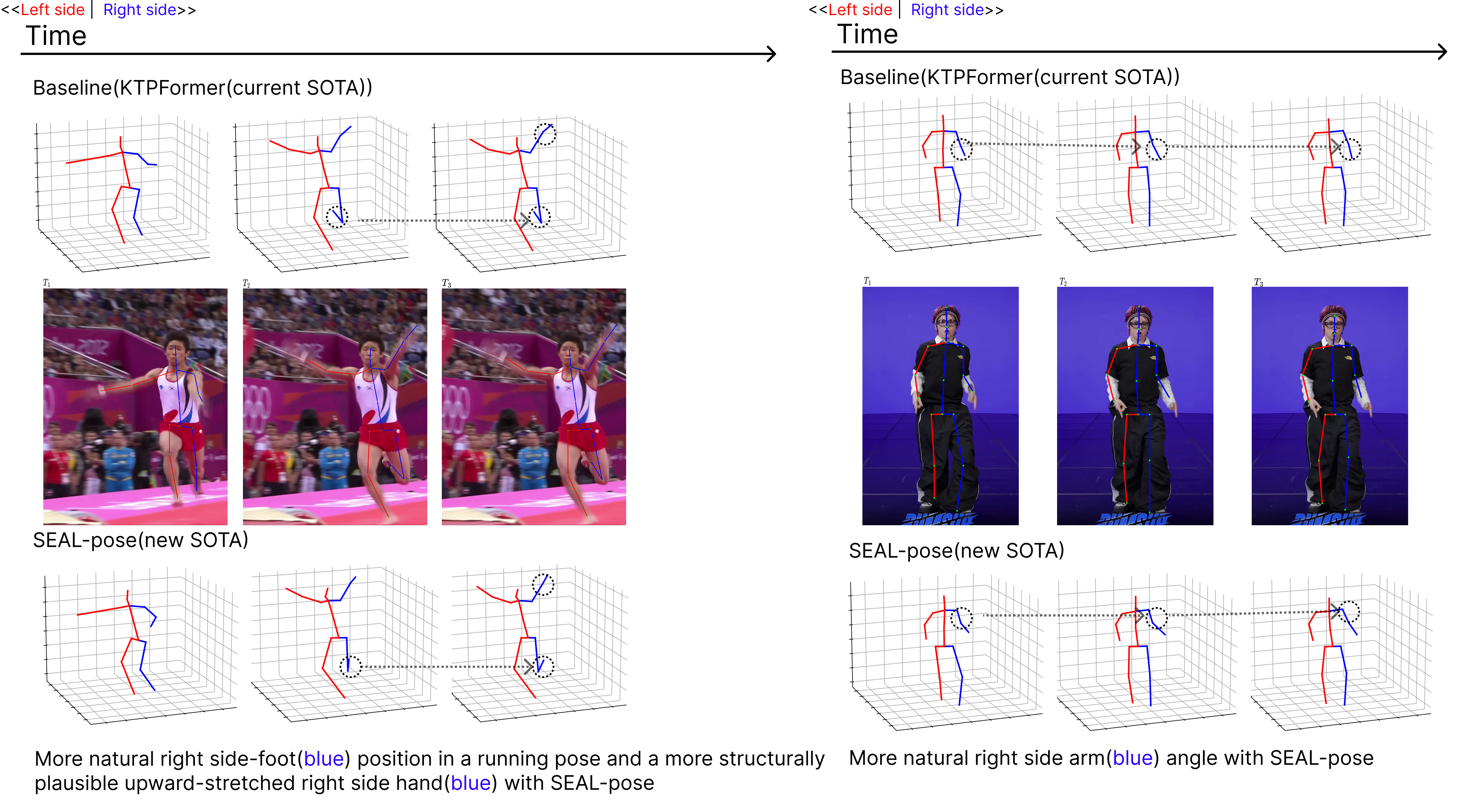}
    \caption{\textbf{Qualitative Comparisons of \sealp~with KTPFormer~\citep{peng2024ktpformer} on an In-The-Wild Images.} Provides visualizations on unseen In-The-Wild images, illustrating that our method produces structurally plausible poses in out-of-distribution settings.}
    \label{fig:inthewild2}
\end{figure*}

Fig.~\ref{fig:inthewild},~\ref{fig:inthewild2} shows qualitative In-The-Wild comparisons between the baseline(PoseFormerV2, KTPFormer) and \sealp. \sealp~often yields more structurally plausible poses, with more natural limb articulations (e.g., elbow/shoulder flexion) and more consistent left–right configurations (red/blue denote left/right), particularly for challenging poses such as hands-together or asymmetric arm motions. Overall, these visualizations suggest that incorporating the loss-net helps maintain structural consistency under noisy 2D observations.


\subsection{Effect of Input Mechanism on Loss-Net Energy.}
~\label{appendix:rep}For joint $j$, let $\mathbf{k}^{2D}_j$ and $\mathbf{k}^{3D}_j$ denote the 2D and 3D keypoints, respectively.
We define the coordinate vector as $\mathbf{k}^{\text{coord}}_j=[\mathbf{k}^{2D}_j,\mathbf{k}^{3D}_j]$ and use a one-hot joint label vector $\mathbf{e}_j$.
Depending on the mechanism, the loss-net node input is either
$\mathbf{h}_j=[\mathbf{k}^{3D}_j,\mathbf{e}_j]$ (3D-only) or
$\mathbf{h}_j=[\mathbf{k}^{\text{coord}}_j,\mathbf{e}_j]=[\mathbf{k}^{2D}_j,\mathbf{k}^{3D}_j,\mathbf{e}_j]$ (2D--3D fused with joint labels).
We evaluate four mechanisms to represent 2D/3D information for learning a 2D--3D compatibility energy.

\hypertarget{mdef:1}{}\textbf{M1 (3D-only prior)} uses 3D-only inputs $\mathbf{h}_j=[\mathbf{k}^{3D}_j,\mathbf{e}_j]$ to compute a graph energy.
\hypertarget{mdef:2}{}\textbf{M2 (bilinear coupling)} forms global descriptors $g_{2d}$ and $g_{3d}$ from $\{\mathbf{k}^{2D}_j,\mathbf{e}_j\}_j$ and $\{\mathbf{k}^{3D}_j,\mathbf{e}_j\}_j$, and couples them via $E=\mathrm{bilinear}(g_{2d},g_{3d})$.
\hypertarget{mdef:3}{}\textbf{M3 (early fusion)} computes a unified graph energy on fused inputs $\mathbf{h}_j=[\mathbf{k}^{2D}_j,\mathbf{k}^{3D}_j,\mathbf{e}_j]$.
\hypertarget{mdef:4}{}\textbf{M4 (SEAL)} uses a decomposed energy $E=E_{\text{local}}(x,y)+E_{\text{global}}(y)$, where $E_{\text{global}}(y)=E(\{\,[\mathbf{k}^{3D}_j,\mathbf{e}_j]\,\}_j)$ and $E_{\text{local}}(x,y)=\sum_{j=1}^{J} (y)^{\top} W_j \tilde{y}$. Compared to early fusion (M3), M4 couples $x$ and $y$ only through the global feature, without joint-aligned 2D--3D node interactions.






Table~\ref{tab:input_design_mpjpe} compares four loss-net input mechanisms (M1--M4) across three backbones.
M3 performs best overall, achieving the lowest MPJPE on VideoPose and SemGCN, while M1 is best on SimpleBaseline.
The MLP-based loss-net shows relatively consistent behavior across mechanisms, with M3 generally performing strongly.
Graph-based loss-nets can represent structural dependencies more effectively by modeling joint interactions and multi-hop relational patterns, which increases their expressive power compared to an MLP-based loss-net. However, this added expressivity also makes performance more sensitive to how 2D and 3D information is injected: in our ablations, we observed slightly more rank swapping across input-mechanism variants under the graph loss-net. This sensitivity highlights that carefully designing the input mechanism is crucial for fully leveraging graph-based reasoning. Empirically, M3 performed best by exposing joint-aligned 2D–3D compatibility cues for local learning and aggregating them via message passing into a globally consistent and plausible pose energy. Overall, our input-mechanism design(M3) is not a minor implementation choice but a necessary and impactful component for 3D HPE, and the early-fusion outcome constitutes a key observation from our design process.

\begin{table}[t]
\centering
\caption{\textbf{MPJPE sensitivity to loss-net input designs (M1--M4).}
MPJPE is reported for four input designs with the backbone baseline.
Best results among M1--M4 are boldfaced; Min--Max (Range) summarizes variation across designs.}
\label{tab:input_design_mpjpe}
\scriptsize
\setlength{\tabcolsep}{3pt}

\resizebox{\columnwidth}{!}{%
\begin{tabular}{@{}lcccccc@{}}
\toprule
Backbone
& Baseline
& \hyperlink{mdef:1}{M1}
& \hyperlink{mdef:2}{M2}
& \hyperlink{mdef:3}{M3}
& \hyperlink{mdef:4}{M4}
& Min--Max (Range) \\
\midrule
SimpleBaseline & 80.9 & \textbf{66.6} & 67.0 & 68.2 & 66.8 & 66.6--68.2 (1.6) \\
SemGCN         & 74.5 & 64.3 & 63.2 & \textbf{62.9} & 67.3 & 62.9--67.3 (4.4) \\
VideoPose      & 66.4 & 66.5 & 66.5 & \textbf{62.2} & 65.9 & 62.2--66.5 (4.3) \\
\midrule
Avg. (3 backbones) & 73.9 & 65.8 & 65.6 & \textbf{64.4} & 66.7 & -- \\
\bottomrule
\end{tabular}%
}
\end{table}

\clearpage
\subsection{Cross-dataset experiment}
\begin{figure}[t]
  \centering

  \begin{subfigure}[t]{\linewidth}
    \centering
    \begin{minipage}[t]{0.49\linewidth}
      \centering
      \includegraphics[width=\linewidth]{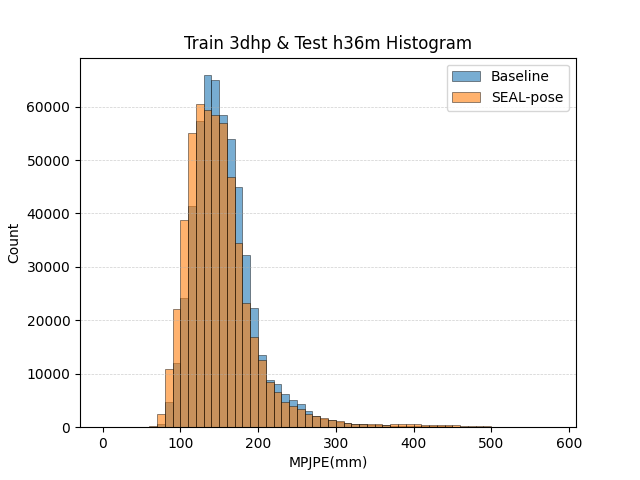}
    \end{minipage}
    \hfill
    \begin{minipage}[t]{0.49\linewidth}
      \centering
      \includegraphics[width=\linewidth]{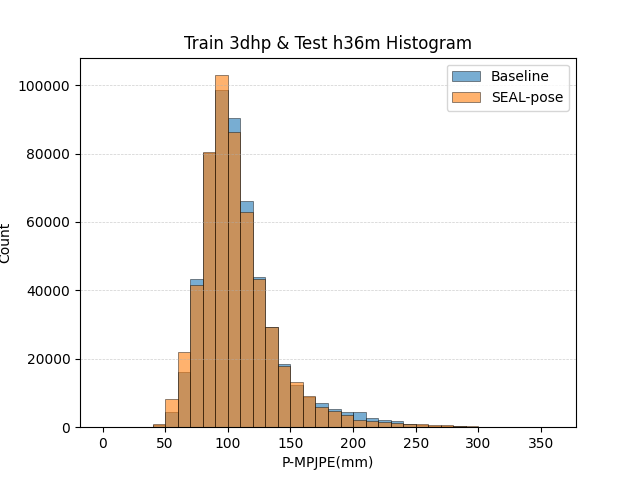}
    \end{minipage}
    \caption{Train 3DHP $\rightarrow$ Test H36M (MPJPE / P-MPJPE).}
    \label{fig:row_train3dhp_testh36m}
  \end{subfigure}

  \vspace{0.6em}

  \begin{subfigure}[t]{\linewidth}
    \centering
    \begin{minipage}[t]{0.49\linewidth}
      \centering
      \includegraphics[width=\linewidth]{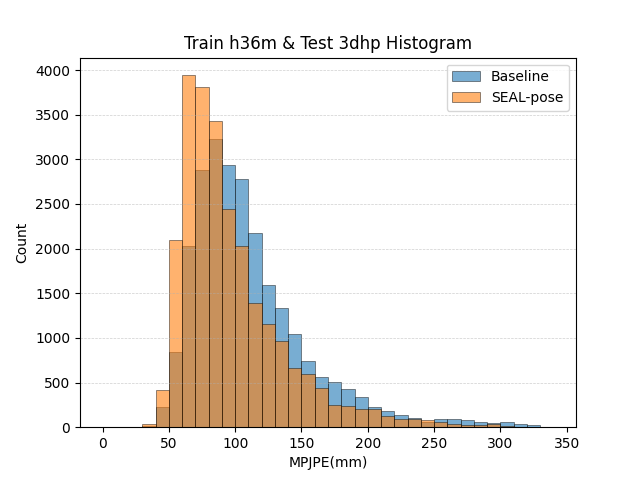}
    \end{minipage}
    \hfill
    \begin{minipage}[t]{0.49\linewidth}
      \centering
      \includegraphics[width=\linewidth]{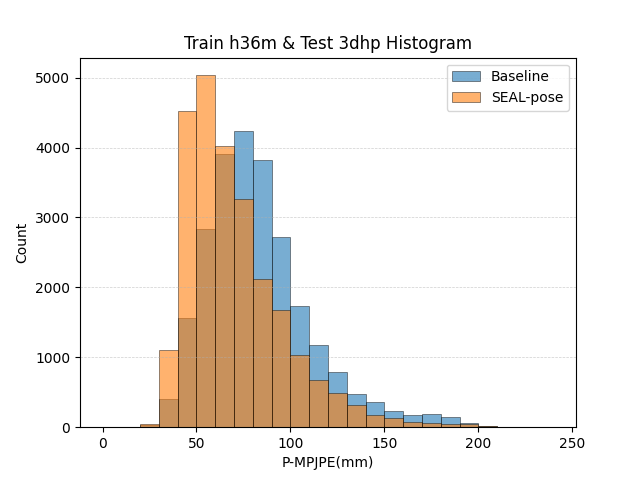}
    \end{minipage}
    \caption{Train H36M $\rightarrow$ Test 3DHP (MPJPE / P-MPJPE).}
    \label{fig:row_trainh36m_test3dhp}
  \end{subfigure}

  \caption{\textbf{Cross-dataset error distributions in PoseFormerV2.} \sealp~ closely follows the baseline distribution and does not introduce additional degradation under dataset shift, while typically reducing the high-error tail.}
  \label{fig:cross_dataset_hists}
\end{figure}

Fig.~\ref{fig:cross_dataset_hists} plots error histograms (MPJPE / P-MPJPE), where each bar counts the number of test samples falling into the same error bin.
Across both transfer directions, SEAL-pose shows a left-shifted distribution compared to the baseline.

\end{document}